\documentclass[journal]{IEEEtran}
\ifCLASSINFOpdf
\else
\fi

\usepackage{graphicx}
\usepackage{amsmath}
\usepackage{amssymb}
\usepackage{stfloats}

\usepackage{xcolor}
\usepackage{url}
\usepackage{tensor}
\usepackage{mathtools}
\usepackage{amsmath}
\usepackage{bbm}
\usepackage{enumerate}
\usepackage[ruled,linesnumbered]{algorithm2e}
\usepackage{siunitx}
\usepackage{tabularx}
\usepackage{multirow}
\newcommand{\tabincell}[2]{\begin{tabular}{@{}#1@{}}#2\end{tabular}}
\usepackage{booktabs}
\usepackage{makecell}

\newcommand{\tabitem}{~~\llap{\textbullet}~~}

\begin{document}
%

\title{Image-Guided Navigation of a Robotic Ultrasound Probe for Autonomous Spinal Sonography Using a Shadow-aware Dual-Agent Framework}
%
%
%

\author{Keyu Li, Yangxin Xu, Jian Wang, Dong Ni, Li Liu$^{*}$, and Max Q.-H. Meng$^{*}$, \textit{Fellow}, \textit{IEEE}
\thanks{This work was partially supported by National Key R\&D program of China with Grant No. 2019YFB1312400, Hong Kong RGC GRF grant \#14210117, Hong Kong RGC TRS grant T42-409/18-R  and Hong Kong RGC GRF grant \#14211420 awarded to Max Q.-H. Meng.}
\thanks{K. Li, Y. Xu and L. Liu are with the Department of Electronic Engineering, The Chinese University of Hong Kong, Hong Kong, China (e-mail: kyli@link.cuhk.edu.hk; yxxu@link.cuhk.edu.hk; liliu@cuhk.edu.hk).}%
\thanks{J. Wang and D. Ni are with the School of Biomedical Engineering, Shenzhen University, Shenzhen, China. (e-mail: wangjian2018@email.szu.edu.cn; nidong@szu.edu.cn).}
\thanks{Max Q.-H. Meng is with the Department of Electronic and Electrical Engineering of the Southern University of Science and Technology in Shenzhen, China, on leave from the Department of Electronic Engineering, The Chinese University of Hong Kong, Hong Kong, and also with the Shenzhen Research Institute of the Chinese University of Hong Kong in Shenzhen, China (e-mail: max.meng@ieee.org).}%
\thanks{$^{*}$Corresponding authors.}%
}

%
%

\markboth{Journal of \LaTeX\ Class Files,~Vol.~14, No.~8, August~2015}%
{Shell \MakeLowercase{\textit{et al.}}: Bare Demo of IEEEtran.cls for IEEE Journals}
%



\maketitle

\begin{abstract}
Ultrasound (US) imaging is commonly used to assist in the diagnosis and interventions of spine diseases, while the standardized US acquisitions performed by manually operating the probe require substantial experience and training of sonographers. 
In this work, we propose a novel dual-agent framework that integrates a reinforcement learning (RL) agent and a deep learning (DL) agent to jointly determine the movement of the US probe based on the real-time US images, in order to mimic the decision-making process of an expert sonographer to achieve autonomous standard view acquisitions in spinal sonography. 
Moreover, inspired by the nature of US propagation and the characteristics of the spinal anatomy, we introduce a view-specific acoustic shadow reward to utilize the shadow information to implicitly guide the navigation of the probe toward different standard views of the spine. 
Our method is validated in both quantitative and qualitative experiments in a simulation environment built with US data acquired from $17$ volunteers. The average navigation accuracy toward different standard views achieves $5.18mm/5.25^\circ$ and $12.87mm/17.49^\circ$ in the intra- and inter-subject settings, respectively. The results demonstrate that our method can effectively interpret the US images and navigate the probe to acquire multiple standard views of the spine.

\end{abstract}


%
\IEEEpeerreviewmaketitle

\section{Introduction}
%
%
%
%

\IEEEPARstart{M}{edical} ultrasound (US) imaging has been widely accepted in a broad range of clinical applications because of its ease of use, non-invasiveness, low cost and real-time capabilities. 
In spinal applications, US imaging is a commonly practiced diagnostic tool for various spine diseases such as scoliosis and low-back pain \cite{rhodes1997review}, and is also frequently used in preprocedural scan \cite{provenzano2013sonographically} and real-time needle guidance during minimally invasive spine procedures \cite{chi2018ultrasound}\cite{harel2016intraoperative}. 

\begin{figure}[t]
\centering
\includegraphics[scale=1.0,angle=0,width=0.48\textwidth]{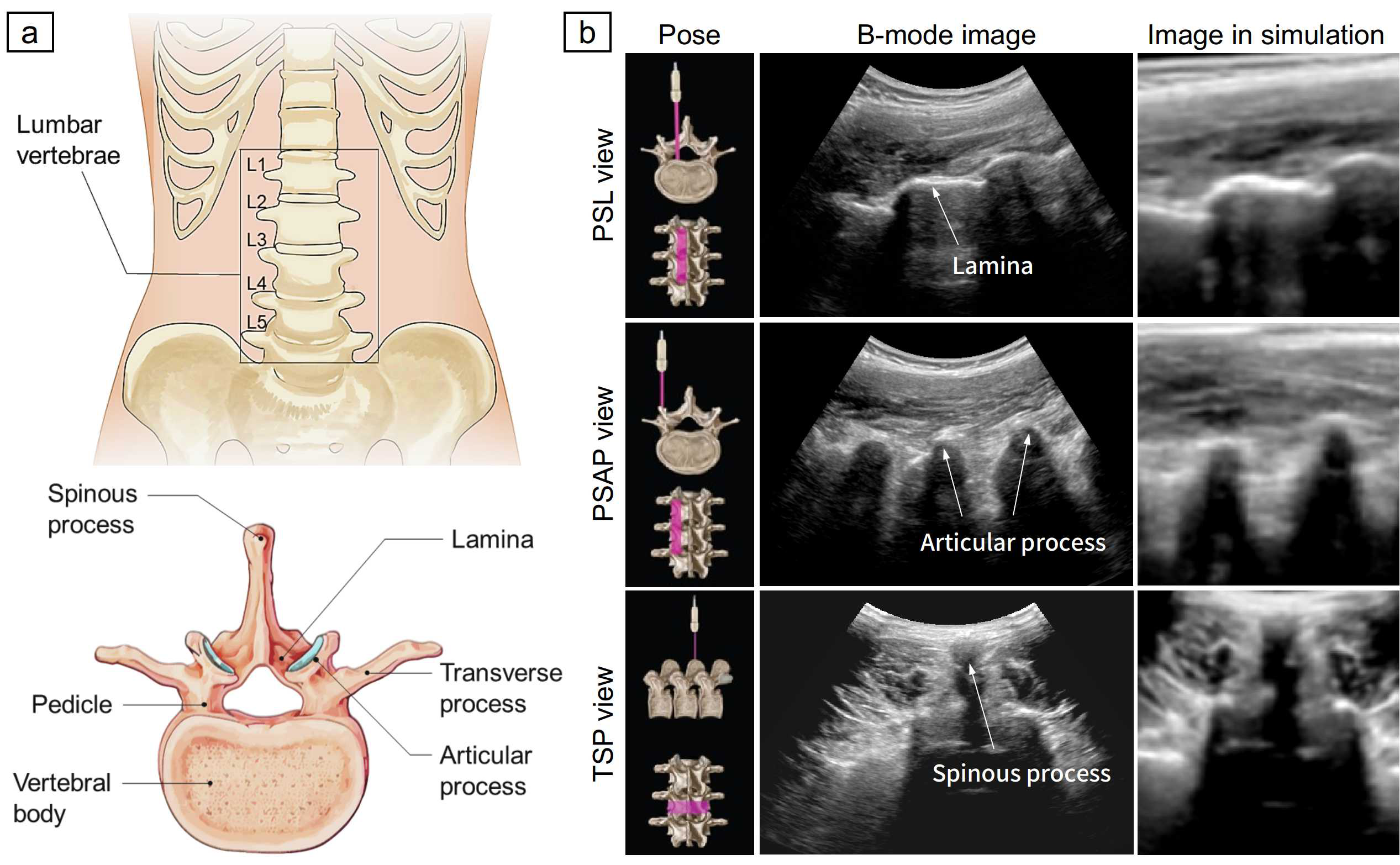}
\caption{(a) Lumbar spine anatomy and (b) US acquisitions of three standard views of the lumbar spine, i.e., PSL: paramedian sagittal lamina view, PSAP: paramededian sagittal articular process view, and TSP: transverse spinal process view of the spine. The left column illustrates the corresponding probe poses \cite{spineUS}. The middle column presents the B-mode images acquired by a clinician from a volunteer. The right column shows the corresponding images acquired with the same probe poses from the virtual patient in our simulation.}
\label{Fig_anatomy}
\vspace{-0.2cm}
\end{figure}

In standardized US acquisitions, the \textit{standard views} are a number of view planes defined by expert consensus to perform US imaging of specific anatomical structures, which usually contain essential information of the anatomy for diagnosis, biometric measurement or interventional guidance \cite{baumgartner2017sononet}\cite{chang2019sonoanatomy}. For example, an overview of the lumbar spine anatomy is show in Fig. \ref{Fig_anatomy}(a). Each lumbar vertebra is composed of a vertebral body, a spinous process, two laminae, two transverse processes, two articular processes and two pedicles \cite{spineUS}. The US examination of the lumbar spine typically involves the imaging of three standard views, namely, the \textit{paramedian sagittal lamina} (PSL) view, \textit{paramedian sagittal articular processes} (PSAP) view, and \textit{transverse spinous process} (TSP) view \cite{karmakar2012sonoanatomy}\cite{ghosh2016ultrasound}, as illustrated in Fig. \ref{Fig_anatomy}(b).
The PSL and PSAP views can be acquired during the paramedian sagittal scan at the level of the lamina and articular process, respectively, and the TSP view is acquired during the transverse scan when the probe is placed over the spinous process. Acquiring these standard views can help the physician identify and locate the spinal anatomical landmarks to perform diagnosis or spine procedures \cite{spineUS}.

However, the current standard view acquisition procedure requires manual navigation of the probe based on the interpretation of the US images and the knowledge of the internal anatomy, which usually requires substantial experience and extensive training of sonographers. Therefore, the sonographers are suffering from heavy physical and cognitive burdens due to excessive workload \cite{disorder2}, and the imaging quality is strongly dependent on the operator \cite{berg2006operator}. In addition, the direct patient contact would increase the risk of infection for frontline medical staff during a pandemic such as COVID-19 \cite{yang2020combating}. In view of this, an autonomous robotic system for US acquisitions holds great promise for relieving user workload, improving imaging results, and reducing the need for direct patient contact \cite{9399640}. 
However, the robot decision-making in medical US acquisitions is a highly challenging task, as it requires the robot to interpret real-time US images and perform visual navigation of the probe, which mimics the decision-making process of an experienced sonographer.

\begin{figure*}[t]
\centering
\includegraphics[scale=1.0,angle=0,width=0.98\textwidth]{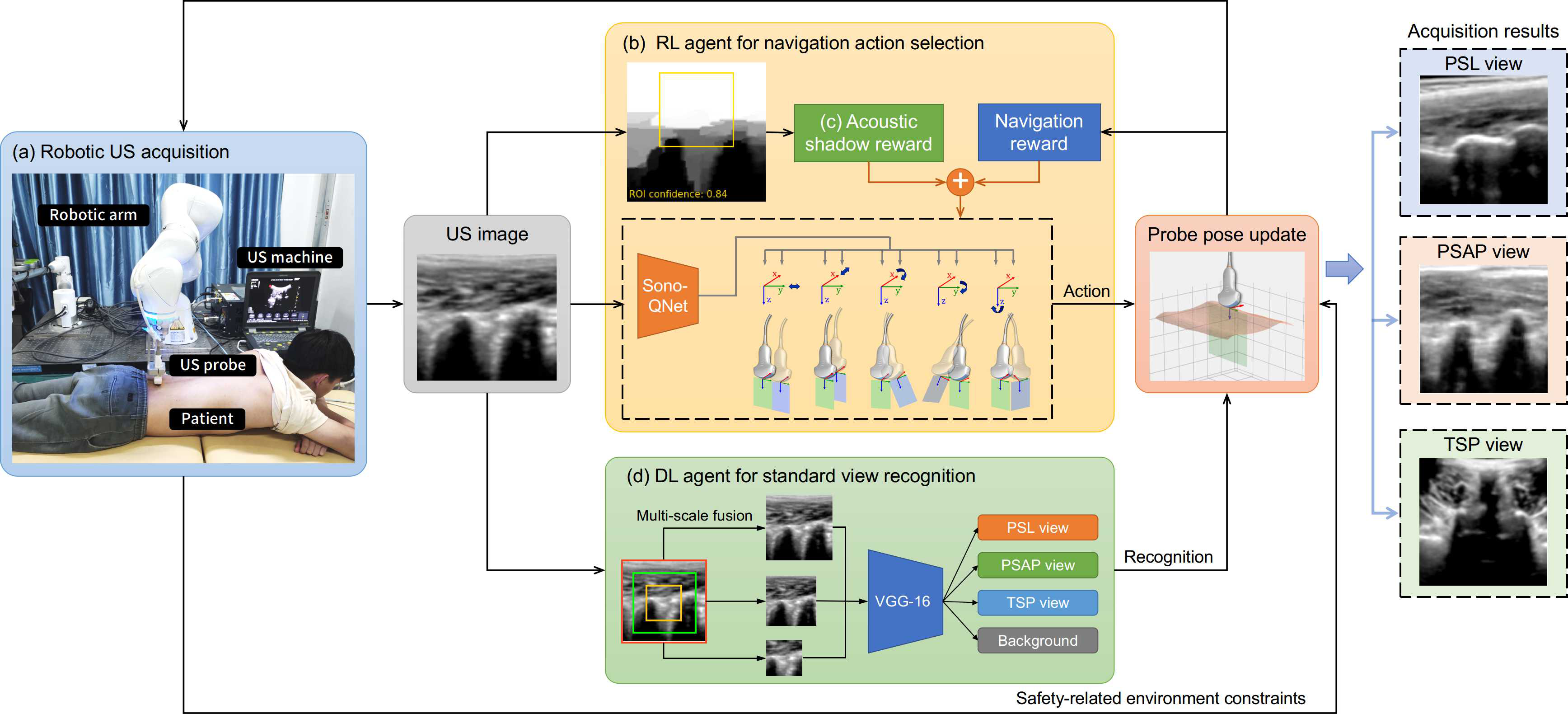}
\caption{Overview of the presented method for autonomous standard view acquisition in robotic spinal sonography. (a) shows the real-world system configuration, where a US probe is controlled by a robotic arm to scan the patient in the prone position. (b) Given the acquired US image as input, the RL agent selects the best navigation action based on the \textit{SonoQNet} to control the 5-DOF movement of the probe. The US confidence map is computed from the US image to calculate the (c) view-specific acoustic shadow reward, which is used in combination with the navigation reward to train the RL agent. Meanwhile, (d) a pre-trained DL agent recognizes the standard views from the US image and jointly determines the movement of the probe under the safety-related environment constraints. The objective of the proposed framework is to automatically acquire three standard views of the lumbar spine (PSL, PSAP and TSP views).}
\label{Fig_method_overview}
\vspace{-0.3cm}
\end{figure*}

As an active field of research in artificial intelligence, reinforcement learning (RL) has become a powerful tool for solving complex sequential decision-making problems in real-world applications \cite{li2019sarl}. Meanwhile, deep learning (DL) has superior ability in learning high-level representation from raw image data and has been intensively studied in US image analysis tasks \cite{aiusanalysis}. 
In this paper, we innovatively introduce a learning-based pipeline to achieve autonomous spinal sonography with a robotic platform, which integrates the RL, DL techniques and a shadow-aware approach to autonomously acquire the standard views of the spine based on real-time US images. An overview of our proposed framework is illustrated in Fig. \ref{Fig_method_overview}. 
The main contributions of this paper are as follows.

\begin{itemize}
\item A deep RL agent is delicately designed and trained end-to-end to plan the 6-DOF movement of a US probe based on real-time US images, in order to autonomously navigate towards the standard views of the lumbar spine.
\item Furthermore, inspired by the nature of US propagation and the characteristics of the spinal anatomy, we introduce a novel view-specific acoustic shadow reward that utilizes the acoustic shadow information to implicitly guide the RL-based navigation.
\item A location-sensitive DL agent is proposed to recognize the standard views of the lumbar spine from real-time US images to provide feedback for the RL agent and jointly determine the movement of the probe through a dual-agent collaborative navigation workflow.
\item A general approach is presented to build a simulation environment for modeling probe-patient interaction in US imaging based on real US data, which can realistically reproduce the real-world US acquisitions and realize continuous state and action spaces for the development of US-guided navigation algorithms.
\end{itemize}

While the work presented in this paper was targeted at the spinal applications, the ideas and methods can also be generalized to the robotic US imaging of other human tissues.

The remainder of this article is organized as follows: Section~II provides an overview of the related work in the fields of US standard view detection and robotic US acquisitions. Then, we introduce the details of our presented method in Section~III, before experiments and results are presented in Section~IV. Conclusions and future perspectives are discussed in Section~V.

\section{Related Work}

\subsection{US Standard View Detection}

As a leading machine learning tool in image analysis, DL has been intensively studied and applied to to the detection of standard views from 2D US image sequences \cite{baumgartner2017sononet} or 3D US volumes \cite{ryou2016automated}. 
Some researchers detect the standard views from 3D US volumes by breaking down the 3D volume into 2D slices for image classification \cite{ryou2016automated}. Lorenz et al. \cite{lorenz2018automated} used anatomical landmark detection to align the 3D volume with a model to localize the standard view planes. 
Other groups applied convolutional neural networks to regress the transformation from the current plane to the standard view plane in 3D US volumes \cite{li2018standard}. However, this kind of prediction may cause abrupt changes in position, which may not be suitable for the robotic control of the US probe. Alansary et al. \cite{alansary2018automatic} customized an RL agent to learn the incremental adjustment of the plane parameters ($ax+by+cz+d=0$) toward the standard views in MRI data, and Dou et al. \cite{dou2019agent} extended this method for standard view detection in 3D US data by performing landmark alignment and RL-based adjustment of the plane parameters. 
However, the aforementioned methods focus on detecting or classifying the standard views from expert-acquired, pre-processed US images and have not taken into account the control of a robotic US probe for autonomous image acquisitions. Instead, we are committed to directly establishing the relationship between the US image content and probe motion control with RL and DL techniques, in order to mimic the decision making process of expert sonographers to realize standard view acquisitions in robotic spinal sonography.

\subsection{Robotic US acquisitions}

A large number of robotic systems have been developed to automate the US imaging of different human tissues. The reader can refer to \cite{9399640} for a literature review.

Different methods have been proposed to plan the movement of the probe during robotic US acquisitions.
In \cite{Hennersperger2016MRI}, the researchers transfer a manually planned scanning path in pre-operative MRI data to the actual patient during the intra-operative stage by registering the patient skin surface extracted from the MRI data to real-time RGB-D data. 
Other researchers directly plan the scanning path on the patient skin surface to cover a region of interest extracted from the RGB-D data \cite{huang2018fully}. Recently, an RL-based method is proposed to control 3 degrees of freedom (DOFs) of a US probe based on the observation of tissue surface by an RGB camera to realize US imaging of a soft target in the presence of occlusion and movement \cite{ning2021autonomic}.
While these surface-based methods have demonstrated the feasibility to keep the probe in contact with the patient and acquire meaningful images, it may not always be feasible to identify the anatomy based on surface landmarks (e.g., in obese subjects) \cite{siddiqui2018ultrasound}, which may reduce the versatility of these methods. Also, the US images that contain rich information of the anatomy have not been fully utilized in the probe motion planning.
Some researchers proposed US-based visual servoing methods \cite{nadeau2014intensity}, but they focus on stabilizing the view of an existing target in the image rather than automatically searching for an anatomy. 
In \cite{nakadate2010implementation}, an automatic scanning strategy is designed to search for the longitudinal plane of the carotid artery by detecting carotid landmarks using well-engineered features. 
In order to develop more generic methods, a number of machine learning-based solutions have been proposed to learn US probe guidance in a data-driven manner. 
Some researchers used imitation learning (IL) algorithms to learn 3-DOF guidance of the probe orientation from expert demonstrations in fetal US imaging \cite{droste2020automatic}. However, the IL-based methods usually rely on complete and accurate demonstrations, which may be intractable or expensive to obtain \cite{hussein2017imitation}.
Other groups applied RL to learn the navigation of a US probe based on self-generated experiences in simulation. Some researchers built the simulation environment with spatially tracked 2D US images acquired by a sonographer on a grid placed on the patient during cardiac \cite{milletari2019straight} and spinal US imaging \cite{hase2020ultrasound}, and used RL to learn probe control with 2 to 4 DOFs. However, these methods have limited state-action spaces due to the design of simulation environments, which would reduce the flexibility of the RL agent. In \cite{li2021autonomous}, we preliminarily developed a deep RL solution to the 6-DOF control of the probe in a simulation environment built with reconstructed 3D volumes that cover the anatomy of interest, which can improve the flexibility of the learned navigation policy.

A distinction between our work and the above works is that we explicitly focus on a system that enables fully automatic 6-DOF control of the US probe based on real-time US images to search for multiple standard views in spinal sonography. Therefore, the system does not rely on manual planning of the scanning path, preoperative tomographic imaging, surface landmark detection or manual initialization of the probe position. Also, our method does not require expert demonstration data, and is not restricted by limited state-action pairs. Instead, we build a simulation environment for probe-patient interaction using real-world US data, and propose a framework based on RL and DL to automatically interpret US images and autonomously plan the movement of the probe.


\section{Methodology}

In this section, we first introduce the simulation environment built with real-world US data to simulate the probe-patient interaction in spinal sonography, which can allow training and testing of RL algorithms for US-guided navigation (Section~III-A). Then, we present the details of our deep RL framework for navigation action selection in Section~III-B. A novel view-specific acoustic shadow reward that utilizes the shadow information to guide the RL-based navigation is introduced in Section~III-C. In Section~III-D, we present a location-sensitive DL agent to recognize the standard views and provide feedback for the RL agent, before we finally introduce the workflow to integrate the RL and DL agents for collaborative navigation in Section~III-E.

\subsection{Simulation of Probe-Patient Interaction}

Since the RL agent learns the optimal policy through trial and error, it is unsafe and impractical to directly train the US probe navigation agent on real patients. A feasible solution is to build a simulation environment with real-world data to train the RL agent. Different from previous studies \cite{milletari2019straight}\cite{hase2020ultrasound} that build the simulation environment with manually acquired, spatially tracked 2D US frames, we build the simulation environment with reconstructed 3D volumes of the lumbar spine acquired in real-world US scans. 
A total of $17$ healthy male volunteers aged $23 \pm 3$ years old were considered in our data acquisition, and three sweeps were performed on each volunteer to acquire the volumetric data that covers the spinal region.
Since we use a reconstructed volume as the virtual patient in our simulation, any slice with arbitrary position and orientation can be sampled in the volume, which can significantly enlarge the state-action spaces for more realistic simulation of probe-patient interaction.

In order to acquire the US data from the volunteers, we built a robotic system as shown in Fig. \ref{Fig_method_overview}(a). The system used a robotic arm (KUKA LBR iiwa 7 R800, KUKA Roboter GmbH, Germany) to control the movement of a convex US probe (C5-1B, Shenzhen Wisonic Medical Technology Co., Ltd, China) attached to its end-effector. 
The volunteers were asked to lie in the prone position on a horizontal bed, and an expert clinician manually configured the US imaging parameters (Depth: $9 cm$, Gain: $55$) and spread the coupling gel over the skin surface of the volunteers. A set of waypoints were manually specified by the clinician to cover the lumbar spine, and the robot automatically moved the US probe to pass through the points under Cartesian impedance control while applying a downward force of $5N$ to ensure acoustic couping. The orientation of the probe was set to make the scan plane parallel to the transverse plane of the body and was kept unchanged during the robotic acquisition. We set a high stiffness ($2000 N/m$) in the horizontal directions and a low stiffness ($50 N/m$) in the vertical direction for compliance. The acquired B-mode images and the corresponding probe poses estimated based on the robotic arm kinematics and US hand-eye calibration \cite{6813647} were recorded for volume reconstruction with a squared distance weighted approach \cite{huang2005development}. After manually screening the resulted $51$ volumes of the $17$ subjects, we removed the data with poor quality and a total of $41$ US volumes were finally used in our simulation. The average volume size of our dataset is $350\times397\times274$ and the size of each voxel is $0.5\times0.5\times0.5mm^3$. After each robotic acquisition, the clinician manually acquired three standard views of the lumbar spine on each volunteer and the corresponding probe poses were recorded, which are considered the ground truth in our experiments.

\begin{figure}[t]
\centering
\includegraphics[scale=1.0,angle=0,width=0.48\textwidth]{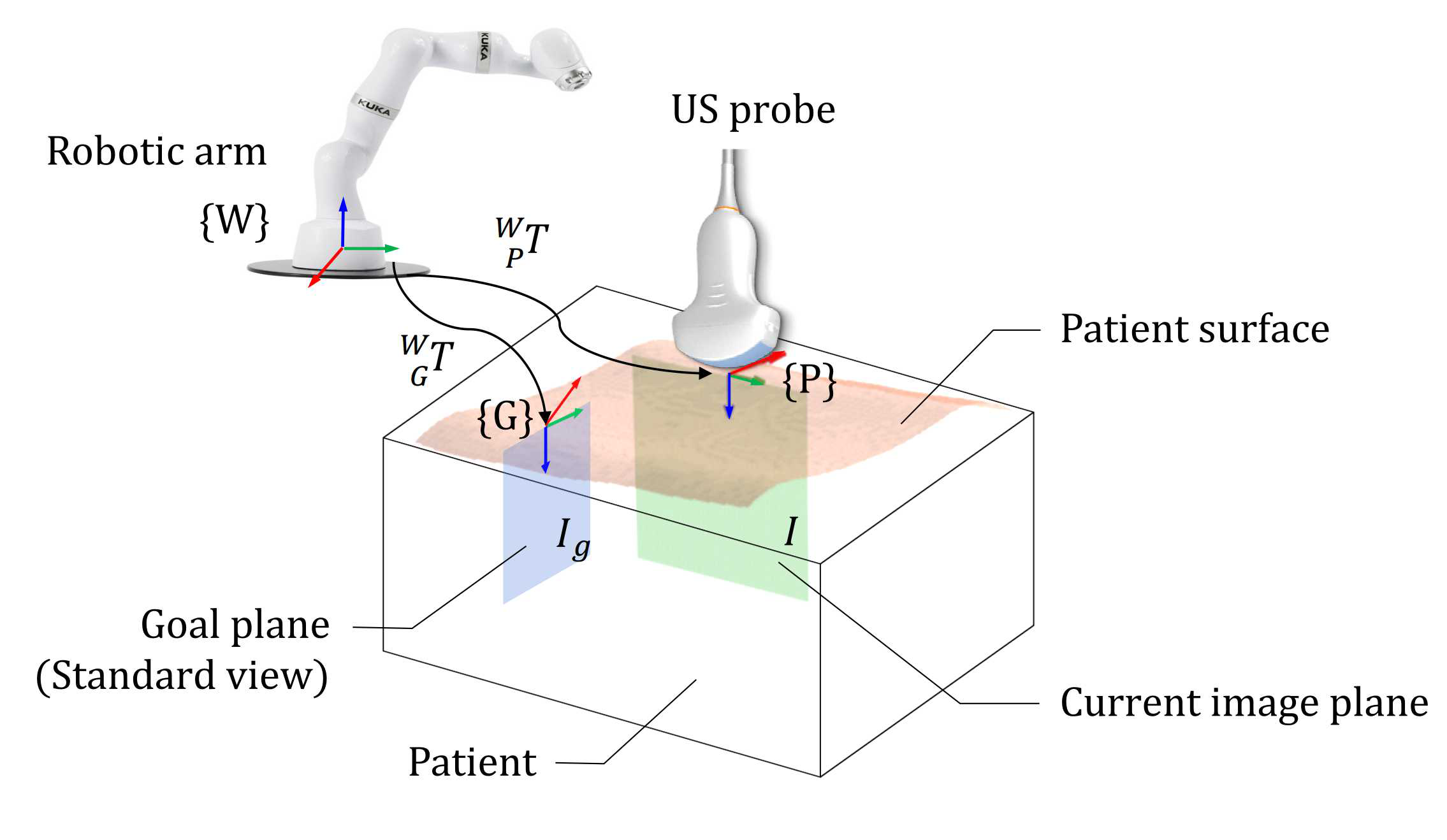}
\caption{Illustration of the simulation environment for robotic spinal sonography. The virtual patients in our simulation are reconstructed 3D volumes of the lumbar spine, and the virtual probe is modeled as a commonly used 2D probe with a square field-of-view. The imaging plane of the probe is set as the $y$-$z$ plane of the probe frame $\{P\}$. The world frame $\{W\}$ is attached to the robot base. The current probe pose and the goal probe pose associated with the target standard view are represented by the transformations $\prescript{W}{P}{\mathbf{T}}$ and $\prescript{W}{G}{\mathbf{T}}$, and the corresponding US images are denoted by $I$ and $I_g$, respectively. }
\label{Fig_simulation}
\vspace{-0.3cm}  
\end{figure}

An overview of our simulation environment is shown in Fig.~\ref{Fig_simulation}. The transformation from the world frame (robot base) $\{W\}$ to the probe frame $\{P\}$ is denoted as $\prescript{W}{P}{\mathbf{T}}$, which is a $4 \times 4$ matrix that describes the current 6-DOF pose of the probe. The goal probe pose $\{G\}$ associated with the target standard view is described as transformation $\prescript{W}{G}{\mathbf{T}}$. Without loss of generality, we assume the virtual probe in our simulation as a commonly used 2D probe with a field-of-view of $150\times150$ pixels, and the imaging plane is defined as the $y$-$z$ plane of the probe frame $\{P\}$. Given a virtual patient (US volume) and an arbitrary probe pose $\prescript{W}{P}{\mathbf{T}}$, the acquired US image $I$ can be uniquely determined by sampling in the volume. In our simulation environment, the skin surface of each virtual patient is extracted as $z=f(x,y)$ based on the intensity of the 3D US volume. In real-world applications, the patient skin surface can be extracted using external sensor data, such as RGB-D or lidar information \cite{Hennersperger2016MRI}.

 The ground truth B-mode images of the three standard views acquired by the clinician and the corresponding images acquired by the virtual probe with the recorded poses in our simulation are compared in Fig. \ref{Fig_anatomy}(b). It can be seen that in spite of a slight deterioration in resolution, the images acquired from the virtual patient can preserve important anatomical structures in the B-mode images, which shows that our simulation environment can realistically reproduce the real-world US acquisitions.

\subsection{Reinforcement Learning for Navigation Action Selection}
\subsubsection{Problem formulation}

In the RL framework, an agent learns the behavior policy by interacting with the environment through a sequence of states, actions and rewards. Here, we formulate the task of the US acquisition robot to observe US images and navigate the probe toward a target standard view as a partially observed sequential decision making problem in the RL framework.

\textbf{States and observations.}
The full state of the probe-holding agent can be represented as $(\prescript{W}{P}{\mathbf{T}},\prescript{W}{G}{\mathbf{T}}, I, I_g)$, indicating the current and goal probe poses and the corresponding US images, as described in Section~III-A. We assume that the patient anatomy and the relative pose between the robot and the patient are unknown. Therefore, the goal probe pose and image are unobservable to the agent, and the agent can only observe the acquired US image $I$ and make navigation decisions. We use a sequence of $4$ consecutive US frames acquired by the agent recently as the observation at time $t$: $s_t=[I_{t-3},\cdots, I_{t}]$ to make use of the dynamic information, similar to \cite{mnih2015human}.

\textbf{Navigation actions.} 
The policy of the RL agent $\pi\colon s_t \mapsto a_t$ maps from the current observation $s_t$ to a navigation action $a_t$. In order to make the learned actions more versatile and independent of the actual patient position and orientation, different from previous methods that represent the actions in the world frame \cite{milletari2019straight}\cite{hase2020ultrasound}, we follow a probe-centric parameterization and define the navigation action $a$ as a transformation operator $\prescript{P}{}{\mathbf{T}}$ with respect to the probe frame $\{P\}$, so that the probe is moved by
\begin{equation}
\prescript{W}{P}{\mathbf{T}} \leftarrow \prescript{W}{P}{\mathbf{T}} \cdot \prescript{P}{}{\mathbf{T}}
\label{basic_pose_update}
\end{equation}

Since $\prescript{P}{}{\mathbf{T}}$ contains $16$ parameters under constraints to represent a valid transformation, directly learning these parameters will be intractable and the interpretability of the learned policy will be limited. Also, a small change of the parameters may result in an abrupt change of the probe pose, which is not favorable in the US probe navigation task. To address these issues, we discretize the action space into $10$ action primitives associated with $5$ DOFs of the probe, as shown in Fig. \ref{Fig_method_overview} (b). Four actions represent the translation of the probe along $\pm x$, $\pm y$ axes of the probe, and six actions represent the rotation around $\pm x$, $\pm y$, $\pm z$ axes of the probe. 
We adopt hierarchical action steps to navigate the probe in a coarse-to-fine manner, similar to \cite{li2021autonomous}. Specifically, the action step $d_{step}$ and $\theta_{step}$ are initialized as $5mm/5^\circ$. $30$ most recent probe poses are stored in a buffer with the format of $(\boldsymbol{p}_{t},\boldsymbol{q}_{t})$, where $\boldsymbol{p}_t$ is the position of the probe at time $t$ and $\boldsymbol{q}_t$ is the quaternion representation of the probe orientation. If at least $3$ pairs among the $30$ most recent probe poses have a pairwise Euclidean distance smaller than $0.01$, the probe pose is assumed to have converged and the action step will be decreased by $1mm/1^\circ$.

\textbf{State transition under constraints.}
Different from previous work that directly apply the navigation action selected by the agent \cite{milletari2019straight}\cite{hase2020ultrasound}, we take into account some safety-related environment constraints in real-world US scans to update the probe pose, as shown in Fig. \ref{Fig_method_overview}. 
First, we consider the practical requirement that the probe be placed over the skin surface to ensure good acoustic coupling. Since the virtual patient in our simulation is roughly parallel with the horizontal plane (see Section~III-A), we use $1$-DOF translational movement of the probe in the $\pm z$ direction to follow the patient surface $z=f(x,y)$.
Second, to ensure good probe-patient contact and guarantee patient safety, the tilt angle of the probe (angle between the imaging plane and the vertical direction) should be limited. Therefore, after the new probe orientation is calculated, the tilt angle is calculated as 
\begin{equation}
\alpha=\arccos (\hat{\boldsymbol{z}}_p,[0,0,-1]^{T})=\arccos [-\prescript{W}{P}{\mathbf{T}}(3,3)]
\end{equation}
\noindent where $\hat{\boldsymbol{z}}_p$ is a unit vector along the $z$-axis of the probe. 
We limit the tilt angle to be smaller than $30^\circ$. If $\alpha > 30^\circ$, the probe orientation will not be updated. 

After the new probe pose is determined under the above constraints, a new US image can be acquired from the virtual patient, and the state of the probe will be updated.

\textbf{Reward.}
In our task, the reward function used for RL training should encourage the agent to minimize the distance to goal, which can be represented by
\begin{equation}
\label{distance}
d_t = \left\|\boldsymbol{p}_t-\boldsymbol{p}_g\right\|_2, \text{\ }
\theta_t = 2 \arccos (| \langle \boldsymbol{q}_t, \boldsymbol{q}_g \rangle |)
\end{equation}
where $\left\| \cdot \right\|_2$ is the L2 norm and $\langle \cdot, \cdot \rangle$ is the inner product. By definition, $d_t$ measures the Euclidean distance between the current probe position and the goal position at time $t$, and $\theta_t$ is the minimum angle required to rotate from the current probe orientation to the goal orientation. 

Some methods construct a dense reward function by classifying the actions as ``good" (moving closer to the goal) or ``bad" (moving away from the goal), and assigning rewards with manually set values \cite{milletari2019straight}\cite{hase2020ultrasound}. Similar to \cite{li2021autonomous}, we design the navigation reward at time $t$ to be proportional to the amount of pose improvement normalized by the action steps

\begin{equation}
\label{r_nav}
\begin{aligned}
r_{nav, t} &= {\Delta d}_t + {\Delta \theta}_t ,\\
\text{where \  }{\Delta d}_t &= \frac{d_{t-1} - d_{t}}{d_{step}} \in [-1,1],\\
{\Delta \theta}_t &= \frac{\theta_{t-1} - \theta_{t}}{\theta_{step}}  \in [-1,1]
\end{aligned}
\end{equation}

\noindent Note that $r_{nav,t} \in [-1,1]$ since the action in each step is either translation or rotation.
In addition, we assign a high reward ($+10$) for task accomplishment ($d_t \leq 1 mm \  \text{and } \theta_t \leq 1 ^\circ$) and add some penalties based on the safety-related constraints. When the tilt angle of the probe $\alpha$ exceeds $30^\circ$, the agent will receive a penalty of $-0.5$. When the probe moves outside the patient volume (the proportion of pixels with non-zero gray value in $I_t$ is less than $30\%$), the agent will get a reward of $-1$. In summary, the reward at time step $t$ is defined as
\begin{equation}
\label{reward}
r_t =
  \begin{cases}
  -1, & \text{if moving out of patient};\\
  -0.5, & \text{if } \alpha > 30^\circ;\\
  10, & \text{if reaching goal} ; \\
  r_{nav, t}, & \text{otherwise}. \\
  \end{cases}
\end{equation}

\textbf{Termination conditions.}
During training, we define four conditions to terminate the navigation: (i) the goal is reached; (ii) the number of steps exceeds the maximum limit (e.g., $120$); (iii) the action step is decreased to zero; and (iv) the probe moves out of the patient. Only the last three conditions are used during inference, since the true location of the standard plane is assumed unknown to the agent.

\subsubsection{Deep reinforcement learning algorithm}

The learning goal of the RL agent is to maximize the return or discounted cumulative future rewards 
\begin{equation}
G_t=r_{t+1}+\gamma r_{t+2}+\gamma^{2} r_{t+3}+\cdots + \gamma^{T-t-1}r_T
\end{equation}
where $\gamma\in(0,1)$ is a discount factor and $T$ is the time step when the episode is terminated. The state-action value function $Q^{\pi}(s,a)$ is defined as the expected return following policy $\mathbf{\pi}$: $Q^{\pi}(s,a)=\mathbb{E}_{\pi}[G_t|s_t=s,a_t=a]$.
The optimal $Q$-function $Q^{*}(s,a)=\max _{\pi}Q^{\pi}(s,a)$ following any policy $\mathbf{\pi}$ is known to satisfy the Bellman equation 
\begin{equation}
Q^{*}(s,a)=\mathbb{E}_{s'}[r(s,a)+\gamma \max \limits_{a'} Q^{*}(s',a')]
\end{equation}

\begin{figure}[t]
\centering
\includegraphics[scale=1.0,angle=0,width=0.49\textwidth]{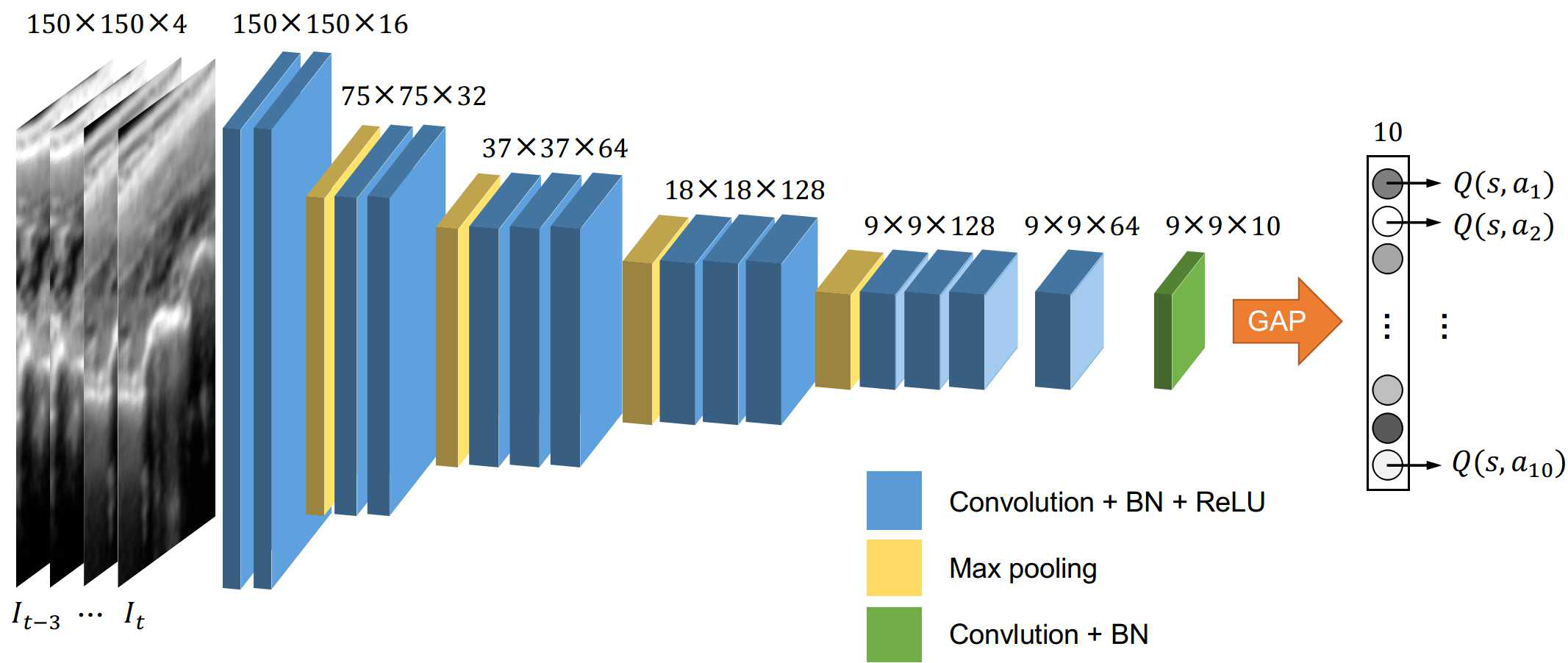}
\caption{Schematic illustration of the SonoQNet architecture for navigation action selection.  The input are $4$ recently acquired US images of size $150\times 150$. The output are the predicted Q-values for the $10$ navigation actions, and the agent will select the action with the highest Q-value. The feature extractor contains $13$ convolutional layers (blue), each followed by batch normalization and ReLU activation. Max pooling (yellow) is performed after the first $4$ convolutional blocks with a filter size of $2\times2$ and a stride of $2$. The size of each feature map is denoted above the blocks. The output of the last convolution+BN block (green) are $10$ class score maps associated with the $10$ navigation actions, which are finally aggregated by global average pooling (GAP) to approximate the Q-values. }
\label{Fig_network}
\vspace{-0.4cm}
\end{figure}

If $Q^{*}(s,a)$ is known, then the optimal policy can be determined by ${\pi}^*=\arg \max _{a}Q^{*}(s,a)$. 
In this work, we use a deep neural network $Q(s,a; w)$ to approximate $Q^{*}(s,a)$, and train the network with the deep Q-learning algorithm \cite{mnih2015human}. The deep Q-network architecture used in our method is referred to as \textit{SonoQNet}, as illustrated in Fig. \ref{Fig_network}. It is modified from the SonoNet-16 architecture \cite{baumgartner2017sononet}, which was originally proposed for real-time classification of standard views in fetal US images. The inputs of \textit{SonoQNet} are $4$ recently acquired US images. The ouputs of the convolutional blocks are $10$ class score maps with a size of $9\times9$, associated with the $10$ navigation actions. To make the network suitable for our target application, we forgo the final softmax layer of SonoNet-16, and the class score maps are aggregated by global average pooling (GAP) \cite{lin2013network} to produce estimation of the Q-values for the $10$ navigation actions.
The optimal network parameters $w^*$ can be learned by iteratively updating $w$ with stochastic gradient descent to minimize the loss function
\begin{equation}
L(w_i)=\mathbb{E}_{s,a,r,s'}[(r+\gamma \max \limits_{a'} Q(s',a';w_{i-1})-Q(s,a;w_i))^2]
\label{loss}
\end{equation}
where $w_i$ is the weight in the $i$-th iteration. 

\subsubsection{Implementation details}
The SonoQNet is trained by the temporal-difference method with experience replay and target network techniques described in \cite{mnih2015human}. In our implementation, we train the network every $10$ interaction steps with a batch size of $32$ using Adam optimizer \cite{kingma2014adam}, and the target nerwork is updated every $1k$ training steps. The discount factor $\gamma$ is $0.9$. The capacity of experience replay memory is $100k$. During initialization, the network is updated for $10k$ iterations with a learning rate of $0.01$ on experiences generated by a supervised policy, which selects the actions to minimize the distance-to-goal. Subsequently, the network is trained for $200k$ iterations on self-generated experiences with an $\varepsilon$-greedy policy. The exploration rate $\varepsilon$ linearly decays from $0.5$ to $0.1$ in the first $100k$ interactions steps and remains unchanged thereafter. The learning rate is set to $0.01$ for the first $40k$ training steps, $0.001$ for the next $40k$ steps, $5\text{e-}4$ for the next $30k$ steps, and $1\text{e-}4$ for the remaining steps.

\subsection{View-specific Acoustic Shadow Reward}

\subsubsection{Acoustic shadow estimation with US confidence map}

\begin{figure}[t]
\setlength{\abovecaptionskip}{0.0cm}  
\centering
\includegraphics[scale=1.0,angle=0,width=0.49\textwidth]{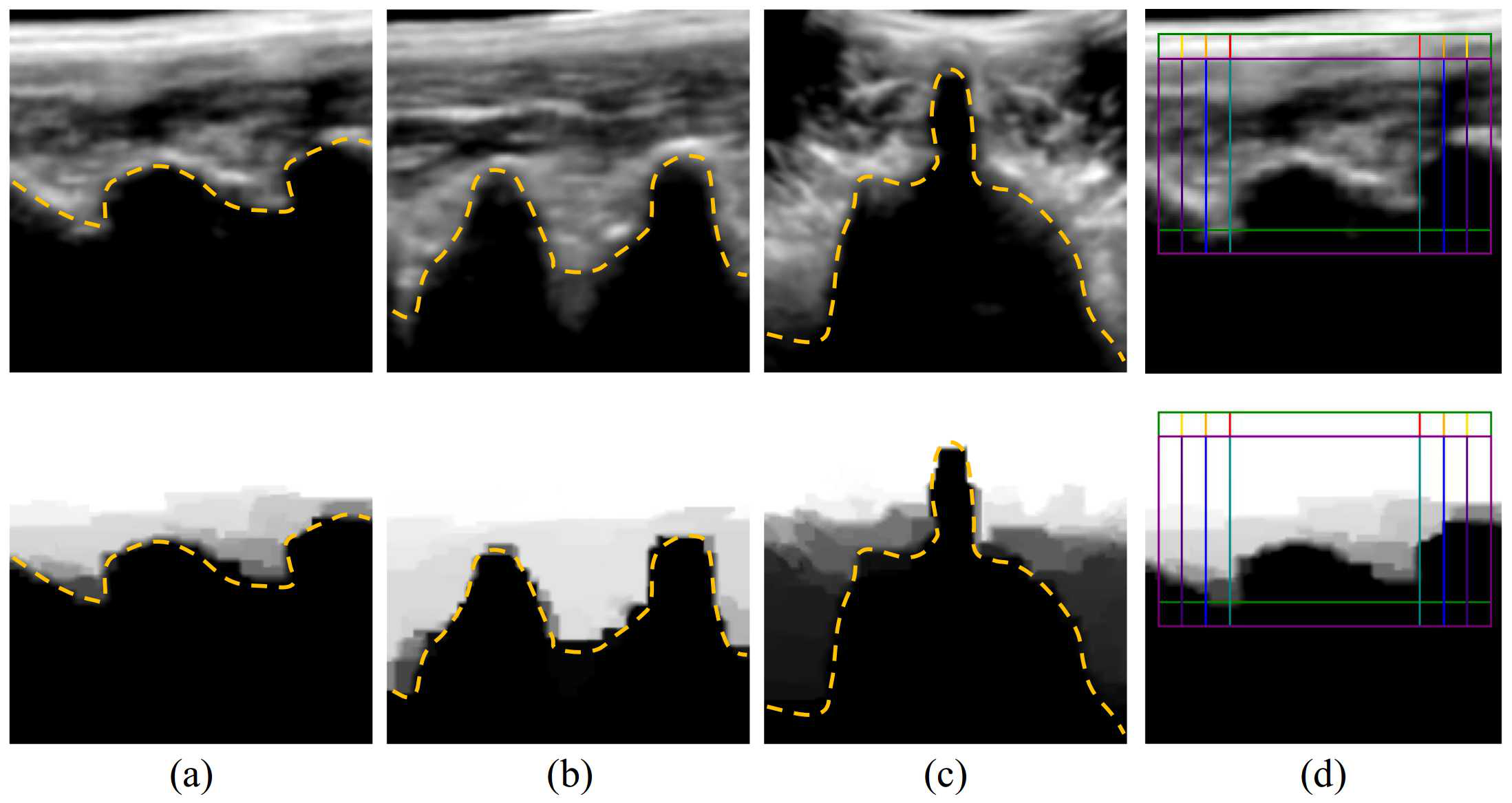}
\caption{(a)-(c) show the US images and the corresponding confidence maps of the PSL view, PSAP view and TSP view of the spine. The acoustic shadow can be seen in the images as area below the yellow dotted line. (d) illustrates the proposed ROI candidates to quantitatively measure the shadow area.}
\label{Fig_shadow}
\vspace{-0.4cm}
\end{figure}

Due to the nature of sound propagation, the US signal will be strongly attenuated at the tissue-bone interface, resulting in acoustic shadows after the interface. In most cases, the shadowing artifacts in US images should be avoided to improve the imaging quality. In spinal sonography, however, the shadows can produce some sonographic patterns that can help locate the spine anatomy \cite{spineUS}. For instance, the shadow patterns in the PSL and PSAP views are referred to as the ``horse head sign" and the ``camel hump sign", and the acoustic shadow in the TSP view appears as a tall dense acoustic shadow, as shown in Fig. \ref{Fig_shadow}(a-c). Therefore, we speculate that additionally taking into account the shadow information may improve the navigation performance in spinal sonography.

In this work, we detect the shadow regions from the US image $I_t$ by calculating its US confidence map at time step $t$, $C_t \leftarrow confidenceMap(I_t)$, $C_t(i,j)\in [0,1]$, based on the method in \cite{confimap} that estimates the per-pixel confidence in the US image to emphasize the uncertainty in shadow regions using a random walks framework. A lower confidence value of a pixel indicates an increased likelihood of acoustic shadow. As shown in Fig. \ref{Fig_shadow}(a)-(c), the possible shadowed regions in the US images are highlighted in the confidence maps. 
Moreover, based on the insight that different standard view acquisition tasks may require different underlying shadow patterns, we calculate the average confidence in a view-specific region of interest (ROI) in the image:

\begin{equation}  
c_t=\frac{1}{|S|}\sum_{(i,j)\in S}{C_t(i,j)}
\label{eq_conf}
\end{equation}
where $S$ denotes the selected ROI for the target standard view. Note that the shadow density in $S$ can be represented by $1-c_t$. The confidence change in the ROI at time step $t$ can be represented by ${\Delta c}_t = c_{t} - c_{t-1} \in [-1,1]$. 

A total of $8$ ROI candidates are proposed, as shown in Fig. \ref{Fig_shadow}(d). They are selected as rectangles centered horizontally in the image and offset $10$ or $20$ pixels from the top edge. The height of the ROI is set as $80$ and the width is chosen from $\{80,100,120,140\}$. The selection of these band-like regions is based on the observation that the top region (with a height of about $10$ pixels) is almost completely located above the tissue-bone interface, and the bottom region (with a height of $50$ pixels) is almost completely located below the tissue-bone interface, which can hardly provide any discriminating shadow information.

\begin{figure}[t]
\centering
\includegraphics[scale=1.0,angle=0,width=0.49\textwidth]{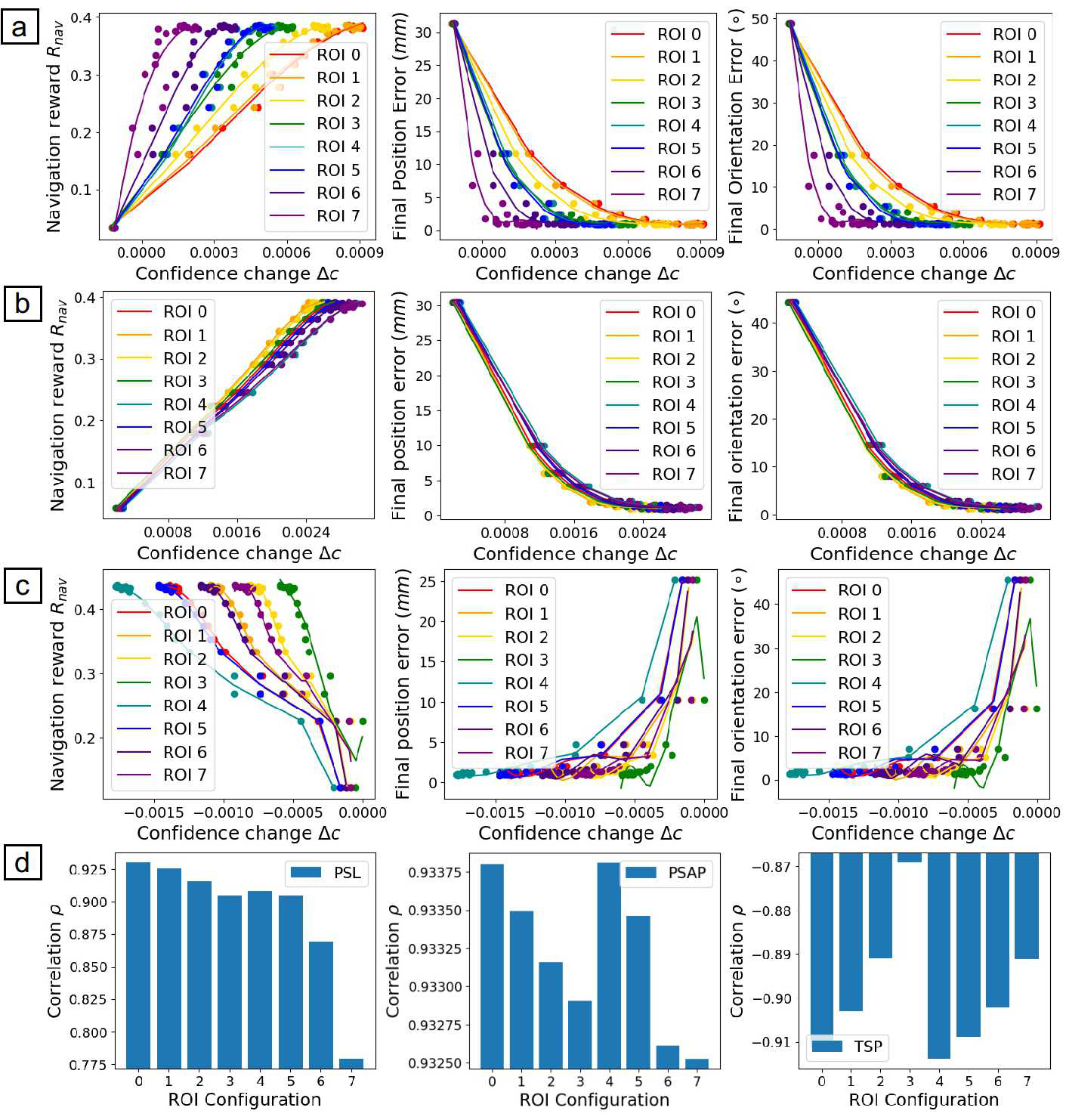}
\caption{(a-c) show the average navigation reward per step $R_{nav}$, final position error $d$, and final orientation error $\theta$ against the average ROI confidence change per step $\Delta c$ during RL training to acquire the PSL, PSAP and TSP views, respectively. The results using different ROI configurations are indicated by different colors. (d) shows the correlation $\rho$ that measures the relationship between the navigation performance and the confidence change in different ROIs during the acquisition of the three standard views.}
\label{Fig_correlation}
\vspace{-0.2cm}
\end{figure}

\subsubsection{Selection of view-specific region of interest}

In order to select view-specific ROIs for different tasks, we analyze the relationship between the navigation performance of the RL agent and the shadow characteristics in each task. As shown in Fig. \ref{Fig_correlation}(a-c), the average navigation reward per step $R_{nav}$, final position error $d$, and final orientation error $\theta$ against the average confidence change per step $\Delta c$ of the RL agent during training in the PSL, PSAP and TSP view acquisitions are illustrated. It can be seen in Fig. \ref{Fig_correlation}(a-b) no matter which ROI configuration is used, as $\Delta c$ increases, the pose improvement becomes greater and the final pose error becomes smaller. This indicates that the navigation performance toward the PSL and PSAP views are positively correlated with an improved confidence in the ROI, and the agent implicitly learns to reduce the shadow area in the ROI during learning of the navigation policy. This is probably because most of the shadow region appears below the ROI in these standard views (see Fig. \ref{Fig_shadow}(a-b)). Differently, for the navigation toward the TSP view, a negative correlation between the navigation performance and the confidence change is observed, which implies that the agent learns to maximize the shadow region during the search for the TSP view, which may be due to the tall dense acoustic shadow in the image center under the spinous process (see Fig. \ref{Fig_shadow}(c)).
To find the ROI configuration that is most related to the navigation performance in each task, we use Pearson's correlation coefficient (PCC) to quantitatively measure the relationship between the navigation performance and the confidence improvement of different ROIs. PCC measures the linear relationship between two variables $X$ and $Y$, which can be calculated by
\begin{equation}
\rho(X,Y)=\frac{\mathbb{E}[(X-\mu_{X})(Y-\mu_{Y})]}{\sigma_{X}\sigma_{Y}}
\label{eq_pearson}
\end{equation}
where $\mu_{X}$, $\mu_{Y}$ are the means of $X$ and $Y$, respectively, and $\sigma_{X}$, $\sigma_{Y}$ are standard deviations of $X$ and $Y$. $\rho(X,Y)\in [-1,+1]$.
In order to take into account both the navigation efficiency and accuracy, we calculate a weighted combination of the PCCs between $R_{nav}$, $d$, $\theta$ and $\Delta c$:
\begin{equation}
\begin{aligned}
& \rho  =\alpha_1 \rho(R_{nav}, \Delta c)+\alpha_2 \rho(-d, \Delta c)+ \alpha_3 \rho(-\theta, \Delta c)\\
\end{aligned}
\label{eq_pearson}
\end{equation}
where the weights are empirically set as $\alpha_1=0.5$, $\alpha_2=\alpha_3=0.25$. Therefore, $\rho\in [-1,+1]$ indicates the correlation between the overall navigation performance and confidence change in the training data. 

The correlation $\rho$ of different ROI configurations in different tasks are shown in Fig. \ref{Fig_correlation}(d). It can be seen that the navigation performance toward the PSL view is most strongly correlated with the confidence improvement in ROI No. 0 , which is a small square region of size $80\times80$, offset $10$ from the top edge. For the acquisition of PSAP view, ROI No. 1 is selected (size $80\times80$, offset $20$ from the top edge). This may be because the tissue-bone interface is deeper in the PSAP view than in the PSL view. For the TSP view acquisition, ROI No. 1 has the largest negative correlation with the navigation performance, which implies that the agent intends to maximize the shadow area in this region.

\subsubsection{Hybrid reward function}

In order to utilize the shadow information to guide the navigation, we introduce a view-specific acoustic shadow reward (ASR) $r_{as,t} = \Delta c_t = c_t - c_{t-1}$ in the original reward function (\ref{reward}) for RL training:
\begin{equation}
\label{new_reward}
r_t =
  \begin{cases}
  -1, & \text{if moving out of patient};\\
  -0.5, & \text{if } \alpha > 30^\circ;\\
  10, & \text{if reaching goal} ; \\
  r_{nav, t}+\lambda r_{sa,t}, & \text{otherwise}. \\
  \end{cases}
\end{equation}
where $\lambda=1$ is used in the acquisition of PSL and PSAP views to encourage confidence improvement in the ROI and $\lambda=-1$ is used for the TSP view to encourage shadow area maximization in the ROI.

\subsection{Deep Learning for Standard View Recognition}
In Section III-B and Section III-C, we have introduced the basic RL agent for US standard view acquisition and its advanced version that takes into consideration the acoustic shadow information to guide the navigation. However, although the RL agent can learn to navigate the US probe based on the acquired US images, the navigation is ``blindly" terminated, i.e., the agent does not know whether it has reached the destination, but simply ends its navigation when the time is up or it has been stuck in a location for a long time. This naive strategy may bring some issues. For example, the agent may be initialized in a suboptimal location at the beginning, and the navigation will be quickly terminated without further exploration. Another possible situation is that the agent may have arrived at the desired standard view, but mistakenly moves away to a suboptimal position. To this end, a better strategy needs to be proposed to enable active termination and exploration of the RL agent. Some researchers proposed to determine the best stopping position of the RL agent based on the Q-values \cite{alansary2018automatic}\cite{dou2019agent}, but these methods may not have a good interpretability. Therefore, we propose to use a pre-trained DL agent to recognize the standard views from the US images to provide feedback for the RL agent and jointly determine the movement of the probe. 

\begin{figure}[t]
      \centering
      \includegraphics[scale=1.0,angle=0,width=0.46\textwidth]{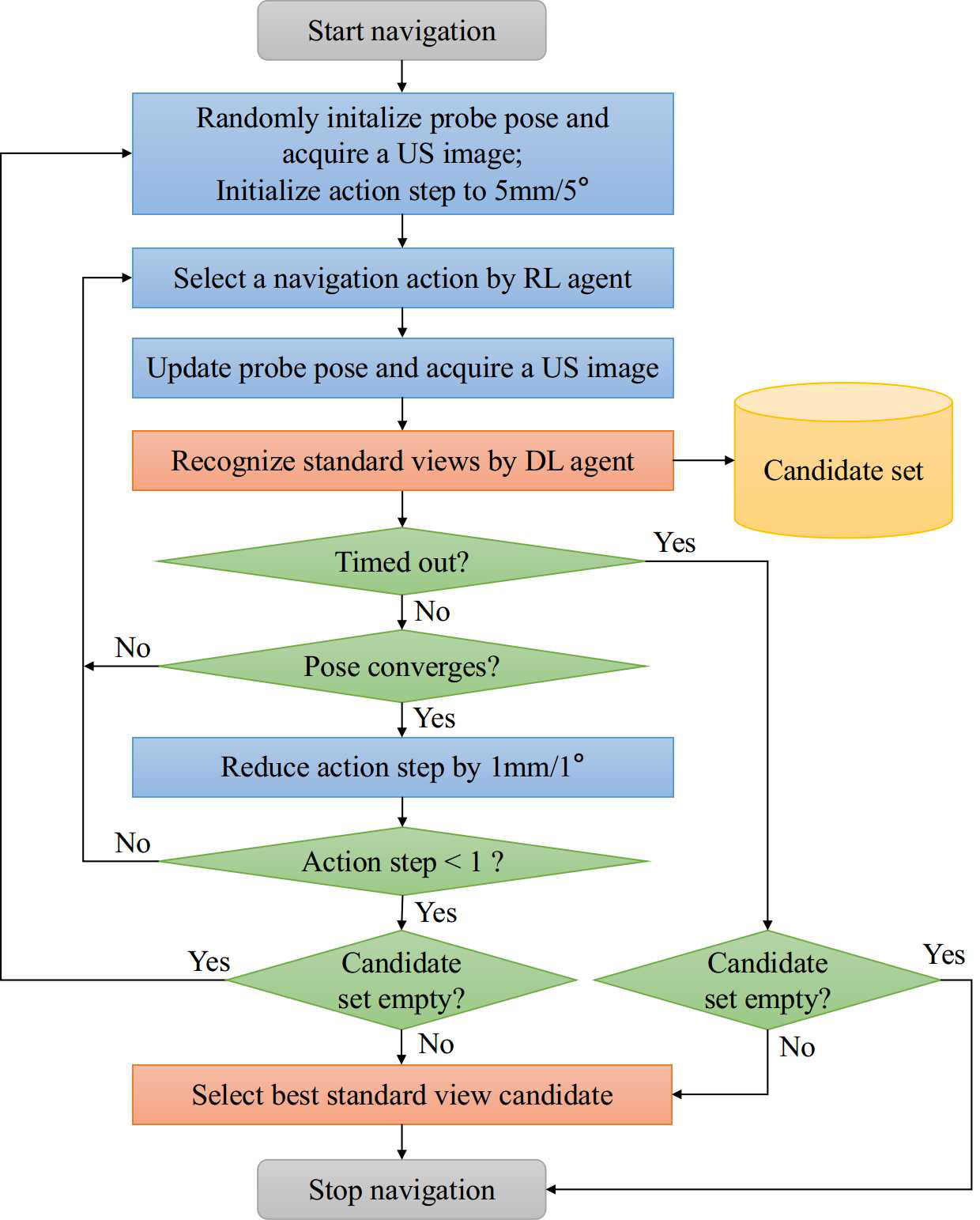}
      \caption{Workflow of the dual-agent collaborative navigation method for standard view acquisition. The blue boxes indicate the operations by the RL agent and the orange boxes indicate the operations by the DL agent. The standard view recognition results of the DL agent are used as feedback to enable active termination and exploration of the RL agent.}
      \label{Fig_dual_agent}
\vspace{-0.4cm}
\end{figure}

In this work, we adopt the VGG-16 \cite{simonyan2014very} as the basic deep neural network model of the DL agent for the standard view recognition task. We initialize the network with publicly available weights trained on the ImageNet dataset \cite{deng2009imagenet}. Then, the original top fully connected layer is replaced with a fully connected layer with $4$ outputs corresponding to the $3$ standard views and background with randomly initialized weights. The network is fine-tuned on our training data (US images with class labels) using stochastic gradient descent and cross-entropy loss with a batch size of $16$ for $50$ epochs to achieve stable performance. A small learning rate of $0.0001$ is used in training. 
Moreover, since the spatial location of pixels in the US image plays an important role to correctly locate the spine anatomy for standard view recognition, we apply a multi-scale fusion (MSF) approach to pre-process the input to make the classification network more sensitive to location features. Multiple scales of the US image (100\%, 75\% and 50\% around the image center) are accumulated as different channels, as shown in Fig. \ref{Fig_method_overview}(d). Note that this approach will not increase the overall complexity of the network.

\subsection{Workflow of Dual-agent Collaborative Navigation}
In order to integrate the RL and DL agents to improve the acquisition results, we propose a workflow to allow collaboration of the two agents to jointly determine the movement of the US probe, as shown in Fig. \ref{Fig_dual_agent}. Each time the RL agent selects a navigation action and acquires an image, the DL agent will detect and save possible standard view images with a predicted probability over $0.5$ in a candidate set. After the RL action step is reduced to zero or time is out, the candidate set will be inspected. If it is not empty, the last recorded standard view candidate during RL navigation will be selected as the best stopping position. We use the last recorded candidate rather than the candidate with the highest predicted probability because i) the standard view recognition result may not be accurate, and ii) it is likely that a position at a later stage in the RL navigation trajectory is more reliable than one in an earlier stage. If time is out but the candidate set is empty, the navigation will be directly terminated. If no standard view is detected during RL navigation and there is still time left, the probe will be randomly repositioned on the patient, and the previous navigation process will be repeated until the total number of steps exceeds the limit.

\section{Experiments}

\begin{table}[tb] \renewcommand\arraystretch{1.25} \footnotesize
\centering
\caption{Summary of Dataset to Train and Test the RL and DL Agents}
\resizebox{0.49\textwidth}{28mm}{
\begin{tabular}{p{0.07\textwidth}<{\raggedright}|p{0.22\textwidth}<{\raggedright}|p{0.18\textwidth}<{\raggedright}}
\Xhline{1pt}
 & RL agent & DL agent \\ 
\Xhline{1pt}
Model & {SonoQNet} & {VGG-16}\\
\hline
\tabincell{l}{Dataset\\ description} &\tabincell{l}{Simulation environments \\(volumetric data and ground- \\ truth poses; see Section III-A)} & \tabincell{l}{ Images with class labels \\(PSL, PSAP, TSP or BG)} \\
\hline
\multirow{4}{*}{Train} & \multirow{4}{*}{\tabincell{l}{25 US volumes acquired from \\ 14 subjects}} &  \multirow{4}{*}{\tabincell{l}{1956 images sampled in \\the 25 training data volumes \\(PSL: 535, PSAP: 381,\\ TSP: 648, BG: 392)}}\\
&&\\
&&\\
&&\\
\hline
\multirow{6}{*}{Test} & \multirow{6}{*}{\tabincell{l}{\tabitem\textit{Intra-subject }setting: \\ 8 unseen data volumes acquired\\ from 3 seen subjects;\\\tabitem\textit{Inter-subject }setting:\\   8 unseen data volumes acquired \\from 3 unseen subjects}} & \multirow{6}{*}{\tabincell{l}{493 images sampled in \\the 8 testing data volumes \\ in the \textit{inter-subject} setting \\(PSL: 118, PSAP: 155, \\TSP: 100, BG: 120)}} \\
&&\\
&&\\
&&\\
&&\\
&&\\
\Xhline{1pt}
\end{tabular}}
\label{T_dataset}
\vspace{-0.2cm}  
\end{table}

\subsection{Dataset}
A summary of the dataset we used to train and test the RL and DL agents is outlined in Table~\ref{T_dataset}.

The RL agent is trained and tested in a total of $41$ simulation environments composed of volumetric US data and ground-truth probe poses, which were collected using the pipeline described in Section~III-A. Among the whole dataset of $41$ US volumes acquired from $17$ subjects, $25$ US volumes of $14$ subjects are used for training, and the remaining data volumes are split into two testing sets associated with two different test settings, i.e., ``\textit{intra-subject}" and ``\textit{inter-subject}", to fully evaluate the effectiveness of the proposed methods. 
In the ``\textit{intra-subject}" setting, the agent is tested on $8$ unseen data volumes acquired from $3$ seen subjects at different times. This setting is designed to evaluate the effectiveness of the method to perform reproducible US acquisitions on familiar subjects in the presence of target displacement and tissue deformation, which is important in some medical applications that require multiple US acquisitions of the same patient, such as pre- and post-operative ultrasonography. 
In the ``\textit{inter-subject}" setting, the agent is tested on $8$ data volumes acquired from $3$ unseen subjects. This task is more challenging as it requires the learned policy to generalize to out-of-distribution data and deal with highly variable patient anatomy. 

In each navigation episode during the training and testing of the RL agent, the pose of the probe is randomly initialized as follows. The horizontal position of the probe is randomly sampled in the center region $\{(x,y):x \sim \mathcal{U}(0.3W,0.7W),y\sim \mathcal{U}(0.2L,0.8L)\}$ of the volumetric data to avoid sampling outside the patient, where $L$, $W$ are the length and width of the US volume, and the height of the probe is then adapted to the patient surface. The $z$-axis of the probe is initialized to be aligned with the $-z$ direction of the world frame, and the probe is randomly rotated around its $z$-axis by $\eta\sim \mathcal{U}(0,360^\circ)\}$. The maximum number of navigation steps in each episode is limited to $120$.

In order to generate a dataset of images with class labels to train and test the DL agent for standard view recognition, we create a training set by sampling in the aforementioned $25$ training data volumes, and a testing set by sampling in the $8$ testing data volumes in the inter-subject setting. Specifically, the position of the probe is sampled with a $10$-pixel interval within $40\%$ around the center of the volume to avoid sampling outside the patient. The orientation of the probe is sampled densely around the standard view orientations by rotating around the $z$-axis by $-10^{\circ}$ to $10^{\circ}$ with an interval of $2^{\circ}$, and sampled sparsely in other orientations by rotating around the $z$-axis by $30^{\circ}$ to $360^{\circ}$ with an interval of $30^{\circ}$. The sampled image is annotated as a standard view image if the probe pose is within $10mm$ and $10^{\circ}$ from the recorded standard view poses and the structural similarity (SSIM) between the acquired image and the ground-truth standard view image is larger than $0.5$. If the sampled probe pose is more than $20mm/20^\circ$ away from all standard views, the sampled image will be annotated as background (BG). Finally, a total of $1956$ images (PSL: $535$, PSAP: $381$, TSP: $648$, BG: $392$) and $493$ images (PSL: $118$, PSAP: $155$, TSP: $100$, BG: $120$) are collected for training and testing purposes, respectively.

\subsection{Evaluation of the Standard View Recognition Module}

\begin{table}[tb] \renewcommand\arraystretch{1.1} \small
\centering
\caption{Classification Scores of Different Models}
\begin{tabular}{lccc}
\toprule
Model&Precision&Recall&F1-score\\
\midrule
VGG-16&	0.9281	&0.9249	&0.9233\\
VGG-16 + MSF	&	\textbf{0.9603}&	\textbf{0.9594}&	\textbf{0.9589}\\
\bottomrule
\end{tabular}
\label{T_classification}
\vspace{-0.2cm}  
\end{table}

\begin{figure}[t]
      \centering
      \setlength{\abovecaptionskip}{0.0cm}
	  \includegraphics[scale=1.0,angle=0,width=0.49\textwidth]{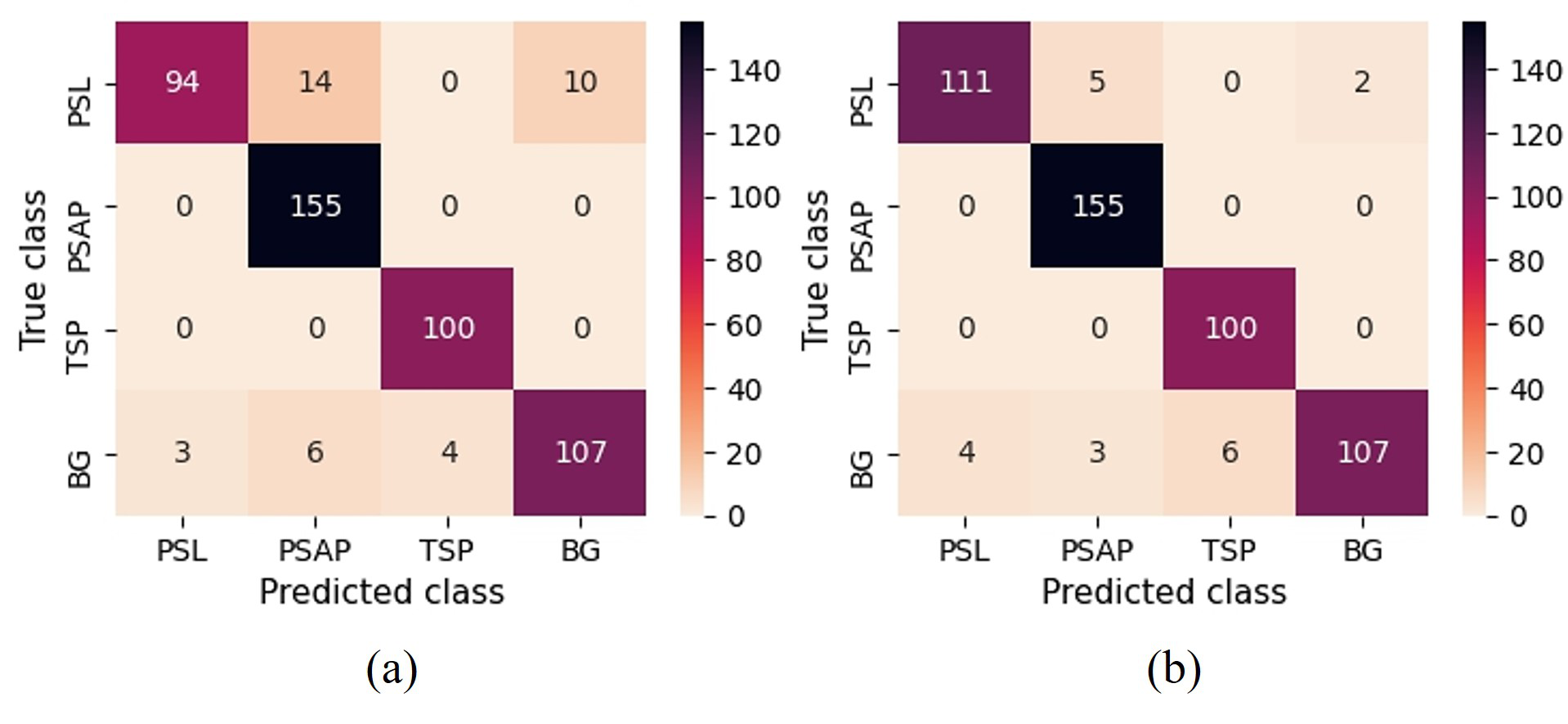}
      \caption{Confusion matrices of the (a) VGG-16 and (b) VGG-16+MSF models for standard view recognition on the test images.}
      \label{Fig_confusion}
\vspace{-0.2cm}  
\end{figure}

We first quantitatively evaluate the classification performance of the DL agent for standard view recognition on the test data using the precision, recall and F1-score. The results of the fine-tuned VGG-16 models with and without using the MSF approach are reported in Table \ref{T_classification}. Moreover, the confusion matrices on the test data are shown in Fig. \ref{Fig_confusion} for a clear comparison. It can be observed that the proposed MSF approach can effectively improve the classification performance of the network without increasing the complexity and number of parameters. As shown in Fig. \ref{Fig_confusion}, all the PSAP and TSP views are classified accurately by both models, while the VGG-16+MSF model can recognize the PSL views with a higher accuracy, showing that our localization sensitive approach can yield better classification performance.

\subsection{Evaluation of the Overall Framework for Standard View Acquisitions}

\begin{table*}[tb] \renewcommand\arraystretch{1.3} \small
\centering
\caption{Performance Evaluation of Different Methods for Standard View Acquisition}
\resizebox{0.99\textwidth}{17mm}{
\begin{tabular}{l|l|cccc|cccc|cccc}
\Xhline{1pt}
\multirow{2}{*}{\tabincell{l}{{Test setting}}} & \multirow{2}{*}{\tabincell{l}{{Method}}} & \multicolumn{4}{c|
}{{Final position error ($mm$)}} & \multicolumn{4}{c|}{{Final orientation error ($^\circ$)}}& \multicolumn{4}{c}{{SSIM}} \\ 
&& PSL view & PSAP view & TSP view& {Average} & PSL view& PSAP view& TSP view& {Average} & PSL view& PSAP view& TSP view& {Average} \\
\Xhline{1pt}
\multirow{4}{*}{\tabincell{l}{Intra-subject}} & RL&9.55$\pm$11.96	&9.06$\pm$15.76	&8.78$\pm$8.35	&9.13$\pm$12.40&	7.27$\pm$13.53	&17.31$\pm$43.38	&6.73$\pm$2.81	&10.44$\pm$26.29 &0.49$\pm$0.18	&0.46$\pm$0.16&	0.51$\pm$0.22	&0.49$\pm$0.33\\
 & RL + ASR & 5.73$\pm$4.89&	7.94$\pm$9.53&	7.14$\pm$10.18&	6.94$\pm$8.53&	6.74$\pm$10.92&	9.57$\pm$22.98	&\textbf{6.19$\pm$3.7}7	&7.50$\pm$14.85 & 0.57$\pm$0.20	&0.44$\pm$0.16	&0.56$\pm$0.20	&0.52$\pm$0.32\\
 & RL + DL & 6.14$\pm$6.41&	6.64$\pm$11.48&	7.18$\pm$6.30&	6.65$\pm$8.42&	4.81$\pm$3.56&	10.02$\pm$28.07&	6.50$\pm$2.99	&7.11$\pm$16.43 &0.57$\pm$0.15	&0.48$\pm$0.14	&0.53$\pm$0.22	&0.53$\pm$0.30\\
 & RL + ASR + DL &\textbf{3.76$\pm$3.42}&	\textbf{4.78$\pm$2.02}&	\textbf{7.01$\pm$9.30}&	\textbf{5.18$\pm$5.84}&	\textbf{4.66$\pm$4.22}&\textbf{	4.51$\pm$2.98}&	6.57$\pm$4.34	&\textbf{5.25$\pm$3.90} &\textbf{0.66$\pm$0.14}	&\textbf{0.48$\pm$0.12}	&\textbf{0.58$\pm$0.18}	&\textbf{0.57$\pm$0.26}\\
\hline
\multirow{4}{*}{\tabincell{l}{Inter-subject}} & RL&24.97$\pm$20.69	&14.05$\pm$12.32	&16.41$\pm$24.63	&18.48$\pm$19.89&	39.96$\pm$55.43	&29.21$\pm$47.56	&19.28$\pm$30.24	&29.48$\pm$45.64 &0.30$\pm$0.09&	0.33$\pm$0.10	&0.46$\pm$0.15	&0.36$\pm$0.20\\
 & RL + ASR & 18.51$\pm$16.37	&14.38$\pm$16.26	&11.71$\pm$22.27	&14.87$\pm$18.51	&\textbf{15.32$\pm$26.08}	&24.67$\pm$46.34	&10.93$\pm$13.61	&\textbf{16.97$\pm$31.69} &0.33$\pm$0.10	&0.33$\pm$0.08&	0.45$\pm$0.16	&0.37$\pm$0.20\\
 & RL + DL & 26.21$\pm$23.02&	11.16$\pm$9.94	&11.07$\pm$8.85	&16.15$\pm$15.35	&27.65$\pm$37.78	&\textbf{19.75$\pm$45.93}&	\textbf{7.20$\pm$4.24}	&18.20$\pm$34.42 &0.39$\pm$0.12	&\textbf{0.37$\pm$0.10}	&0.49$\pm$0.15	&0.42$\pm$0.22\\
 & RL + ASR + DL & \textbf{18.07$\pm$11.35}&	\textbf{10.96$\pm$11.99}	&\textbf{9.57$\pm$10.00}	&\textbf{12.87$\pm$11.14}&	18.39$\pm$31.16	&19.96$\pm$43.29	&14.13$\pm$30.95	&17.49$\pm$35.60 &\textbf{0.43$\pm$0.09}	&0.35$\pm$0.07	&\textbf{0.49$\pm$0.14}	&\textbf{0.43$\pm$0.18}\\
\Xhline{1pt}
\end{tabular}}
\label{T_overall}
\vspace{-0.2cm}  
\end{table*}

Then, we conduct both quantitative and qualitative experiments to compare the performance of our proposed framework with different configurations for the standard view acquisition task. Ablation studies are performed to validate the effectiveness of the three components in our proposed framework: the basic RL agent, the view-specific acoustic shadow reward (ASR), and the dual-agent collaborative navigation. Each of the four variations of our method (i.e., ``RL", ``RL+ASR", ``RL+DL", ``RL+ASR+DL") is evaluated on $24$ random test cases (including $3$ navigation tests on each test data) towards each of the three standard views in both the ``intra-subject" and ``inter-subject" settings. 
The performance is evaluated using the final position and orientation errors, as well as  the structural similarity (SSIM) between the acquired image and the target standard view image, so as to assess the quality of acquisition in terms of both the probe pose and image content. In Table \ref{T_overall} we report the results of all the methods in each of the three standard view acquisition tasks and their averages. A detailed analysis of the results is provided in the following.

\begin{figure}[t]
      \centering
      \setlength{\abovecaptionskip}{0.0cm}
	  \includegraphics[scale=1.0,angle=0,width=0.49\textwidth]{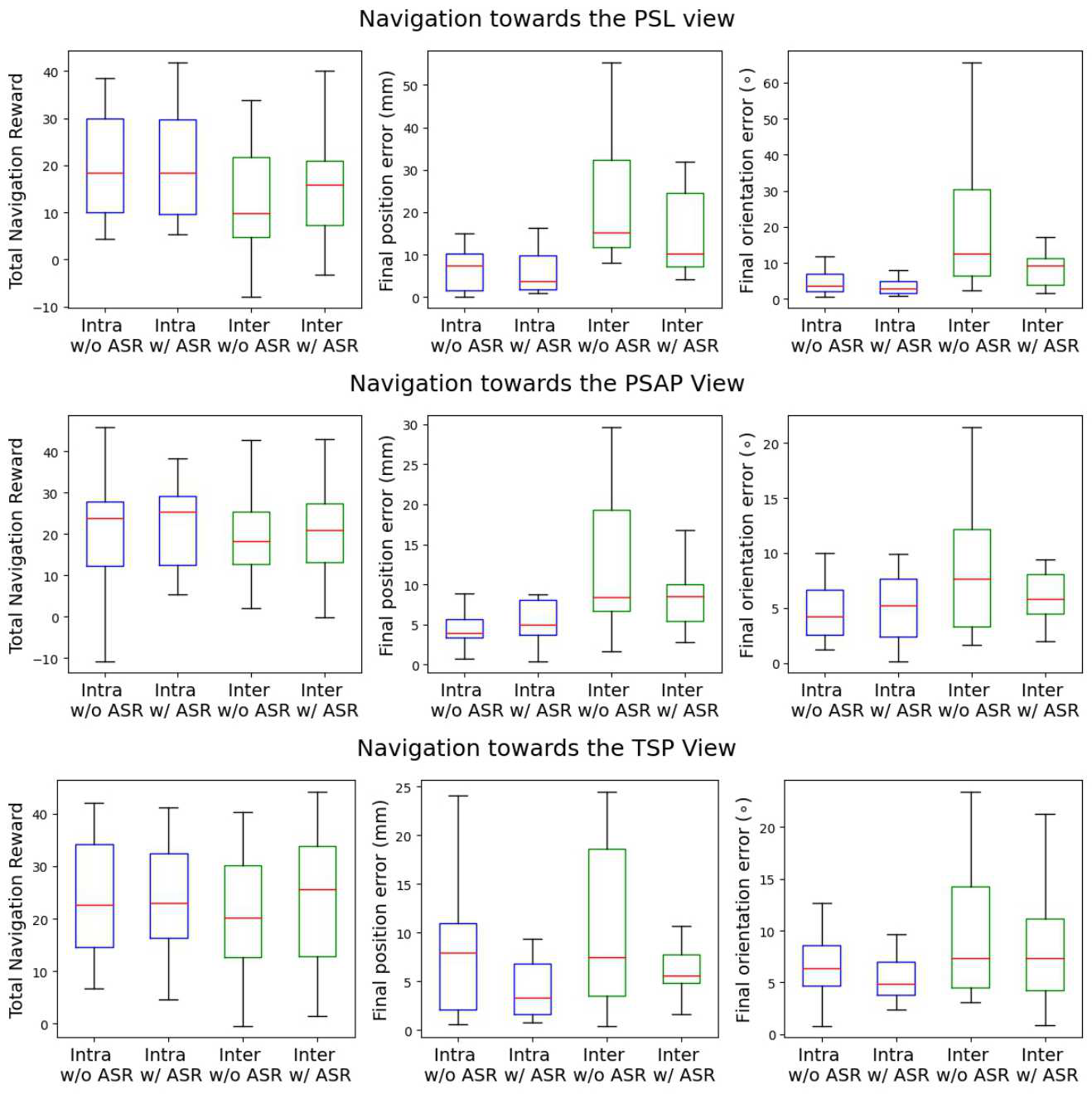}
      \caption{The box-plots illustrate the navigation performance of the RL agents trained with and without using the view-specific acoustic shadow reward (ASR) in the ``intra-subject" (blue) and ``inter-subject" (green) test settings towards the PSL, PSAP and TSP views. The evaluation metrics include the total navigation reward (first column), final position error (second column) and final orientation error (third column). The red lines show the medians.}
      \label{Fig_experiments_asr}
\vspace{-0.2cm}  
\end{figure}

\subsubsection{Evaluation of view-specific acoustic shadow reward}

\begin{figure*}[tb]
\setlength{\abovecaptionskip}{0.1cm}  
\centering
\includegraphics[scale=1.0,angle=0,width=0.8\textwidth]{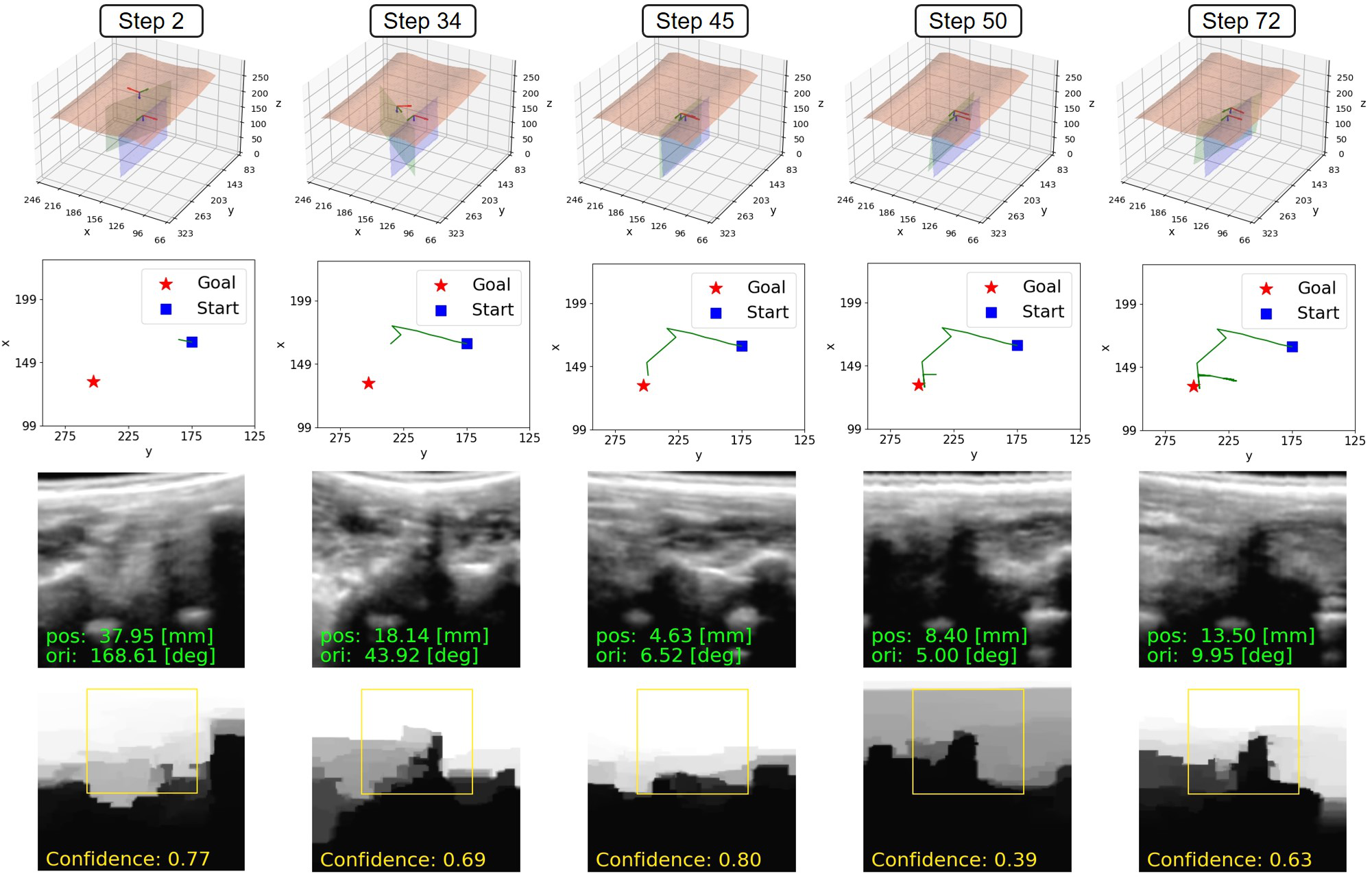}
\caption{Navigation trajectory of the ``RL" agent toward the PSL view in an example test case. The first line shows the 3D plots of the virtual patient surface (salmon), current imaging plane (green), goal plane (blue), and the corresponding probe poses. The second line shows the top-view trajectory of the agent on the horizontal plane (green lines). The start and goal positions are indicated by a blue square and a red star, respectively. The US images are shown in the third line, with the pose errors marked in green. The confidence maps are plotted at the bottom, with the average confidence value of the selected ROI (yellow rectangle) marked in yellow.}
\label{Fig_traj}
\vspace{-0.2cm}  
\end{figure*}

\begin{figure*}[tb]
\setlength{\abovecaptionskip}{0.1cm}  
\centering
\includegraphics[scale=1.0,angle=0,width=0.8\textwidth]{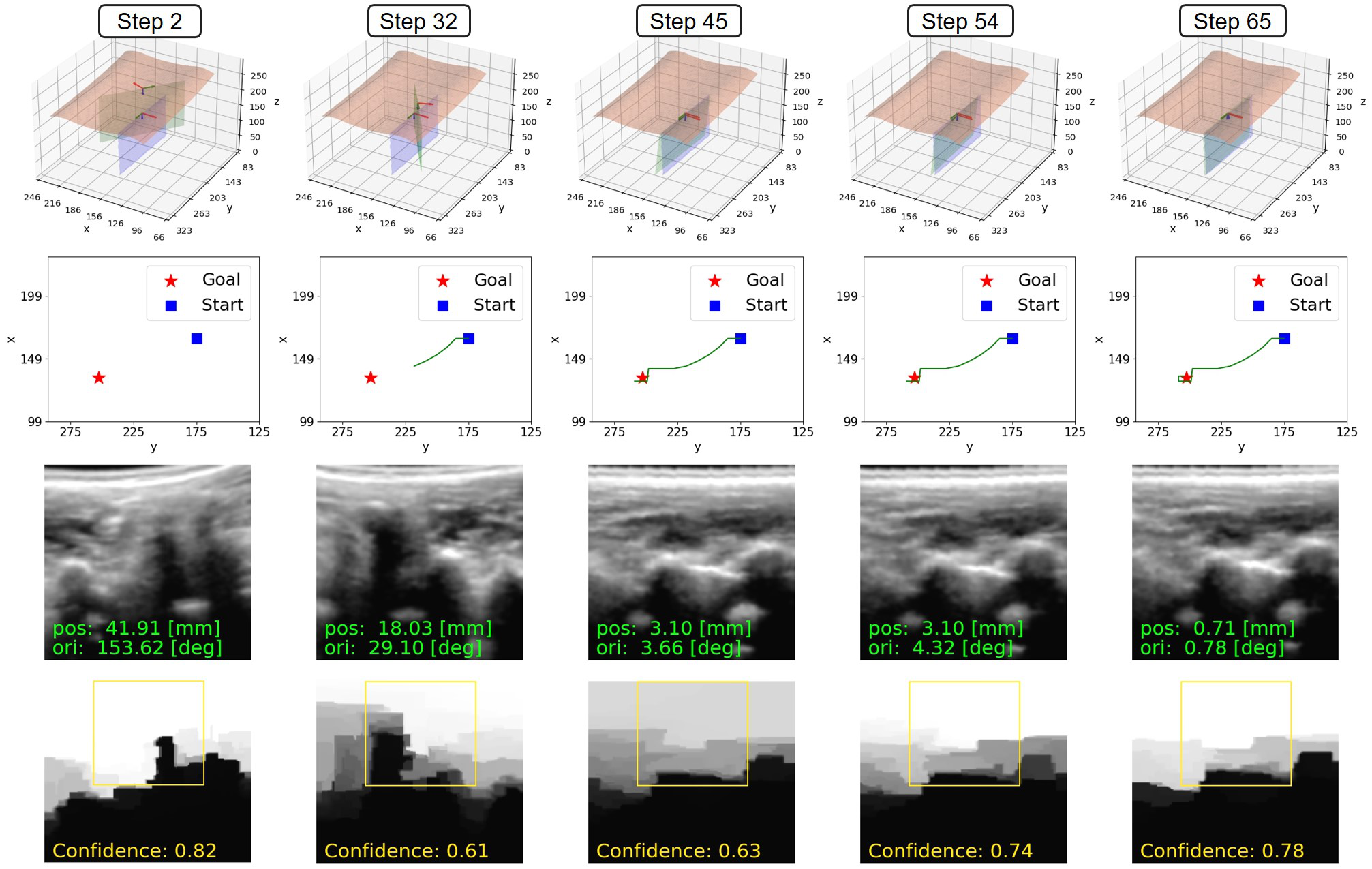}
\caption{Navigation trajectory of the ``RL+ASR" agent toward the PSL view in the same test case as in Fig. \ref{Fig_traj}. The first line shows the 3D plots of the virtual patient surface (salmon), current imaging plane (green), goal plane (blue), and the corresponding probe poses. The second line shows the top-view trajectory of the agent on the horizontal plane (green lines). The start and goal positions are indicated by a blue square and a red star, respectively. The US images are shown in the third line, with the pose errors marked in green. The confidence maps are plotted at the bottom, with the average confidence value of the selected ROI (yellow rectangle) marked in yellow.}
\label{Fig_traj2}
\vspace{-0.2cm}  
\end{figure*}

We first evaluate the RL agents trained with and without using ASR (i.e., by applying the reward function in (\ref{new_reward}) or (\ref{reward})), and compare their navigation performance in the acquisition of the three standard views of the spine. We also compare the cumulative navigation reward $\sum_{t=0}^{T} r_{nav, t}$ and final pose error of the two agents in both test settings in Fig. \ref{Fig_experiments_asr}.
As shown in Table \ref{T_overall} (lines 1-2, 5-6), the navigation accuracy of the ``RL+ASR" agent in both position and orientation shows a remarkable improvement of $2.19mm/2.94^\circ$ and $3.61mm/12.51^\circ$ over the ``RL" agent in the intra- and inter-subject settings, respectively, and the SSIM is also improved by $3\%$ and $1\%$ in the two settings. 
It can be seen in the left column of Fig. \ref{Fig_experiments_asr} that by applying the proposed ASR in the RL training, the cumulative navigation rewards are increased in both test settings in the three standard view acquisition tasks, which shows that using the ASR as an auxiliary reward can implicitly improve the navigation performance of the RL agent. From the second and third columns of Fig. \ref{Fig_experiments_asr}, we can see that the final position and orientation errors are generally lower for the ``RL+ASR" agent compared with the basic ``RL" agent, especially in the more challenging ``inter-subject" setting.

We also compare the navigation performance of the ``RL+DL" and ``RL+ASR+DL" methods to see the effectiveness of the ASR module when using RL-DL collaborative navigation to determine the probe movement. As shown in Table \ref{T_overall} (lines 3-4, 7-8), the ``RL+ASR+DL" method yields a better navigation performance over ``RL+DL", with a final pose error of $5.18mm/5.25^\circ$ and $12.87mm/17.49^\circ$ in the intra- and inter-subject settings, respectively, and the SSIM is increased by $4\%$ and $1\%$ in the two settings. 
The results show that the use of acoustic shadow information in our method can implicitly improve the navigation performance of the RL agent, which is consistent with the prediction in Section~III-C. It should also be noted that a decline in the orientation accuracy by the ``RL+ASR+DL" method toward the TSP view is witnessed, which may be due to the inaccuracy introduced by the DL-based standard view recognition module.

For a qualitative evaluation, we illustrate the navigation trajectories by the ``RL" and ``RL+ASR" agents toward the PSL view in an example test case in Fig. \ref{Fig_traj} and Fig. \ref{Fig_traj2}.
It can be observed that both agents take actions that smoothly navigate the probe toward the target standard view in the first $45$ steps in their navigation. The pose errors of the two agents (marked in green) are greatly reduced to $4.63mm/6.52^\circ$ and $3.10mm/3.66^\circ$, respectively, and the acquired US images show high similarity with the ground truth PSL view (see Fig. \ref{Fig_method_overview}). However, the ``RL" agent mistakenly moves the probe away from the goal in the subsequent $5$ steps, and reaches a location with a low confidence value in the selected ROI (step $50$, $c=0.39$). In the remaining steps, the agent moves even farther away from the goal until the end of its navigation and ``misses the victory". This may be because the agent only learns to minimize the distance-to-goal and cannot interpret the shadow information, so the large acoustic shadows that suddenly appear in the image greatly affect its navigation decisions. 

In contrast, the ``RL+ASR" agent keeps refining the pose of the US probe in a small range after step $45$, concerning both the distance-to-goal and shadow characteristics, and finally stops at a location close to the goal. Meanwhile, it can be observed from the confidence maps in Fig. \ref{Fig_traj2} that the acoustic shadow area in the selected ROI of the ``RL+ASR" agent is gradually decreased during the fine-tuning of the probe in steps $45$ to $65$. Since this shadow-aware agent learns to maximize both the navigation reward and the auxiliary shadow reward, it can utilize the shadow information to guide the navigation for better US acquisition results.

\subsubsection{Evaluation of dual-agent collaborative navigation}

Then, we evaluate the dual-agent collaborative navigation method for the integration of the RL and DL agents, as described in Section~III-E. As shown in Table \ref{T_overall}, ``RL+DL" and ``RL+ASR+DL" denote the methods that use RL-DL collaborative navigation with the RL agent trained without and with ASR, respectively.

We first take a look at the performance of the dual-agent collaborative navigation approach when using the basic ``RL" agent. As shown in Table \ref{T_overall} (lines 1, 3), compared with the ``RL" method, the ``RL+DL" method can significantly reduce the final pose errors and increase the similarity between the acquired images and standard views in the intra-subject setting, especially for the PSL and PSAP view acquisitions. In the inter-subject setting, the final pose error and SSIM of the ``RL+DL" method toward the PSAP and TSP views also see a large improvement over the ``RL" method (see Table \ref{T_overall}, lines 5, 7), while a slight deterioration is found in the final position error during the PSL view acquisition. This may be due to the inaccurate recognition results of the DL agent and the suboptimal navigation trajectory of the RL agent trained without ASR.

Second, we compare the navigation performance of the ``RL+ASR" and ``RL+ASR+DL" methods to validate the effectiveness of the dual-agent collaborative navigation method when used in combination with ASR. As shown in Table~\ref{T_overall} (lines 2, 4, 6, 8), the ``RL+ASR+DL" method further improves over the ``RL+ASR" method in the final position errors toward the standard views in both intra- and inter-subject settings. While the orientation error of the ``RL+ASR+DL" method shows a slight increase by $0.52^\circ$ compared with the ``RL+ASR" method in the inter-subject setting, the final SSIM between the acquired image and the goal image of the ``RL+ASR+DL" method shows a remarkable improvement over ``RL+ASR" by $6\%$. This may be because the dual-agent framework is more sensitive to the US image content since it takes advantage of the standard view recognition results. Since the ultimate objective of our method is to acquire the standard view images that visualize the target anatomical structures, the slight deterioration of the orientation accuracy can be considered as tolerable compared with the significant improvement of image similarity.

\begin{figure}[t]
\setlength{\abovecaptionskip}{0.1cm}  
\centering
\includegraphics[scale=1.0,angle=0,width=0.48\textwidth]{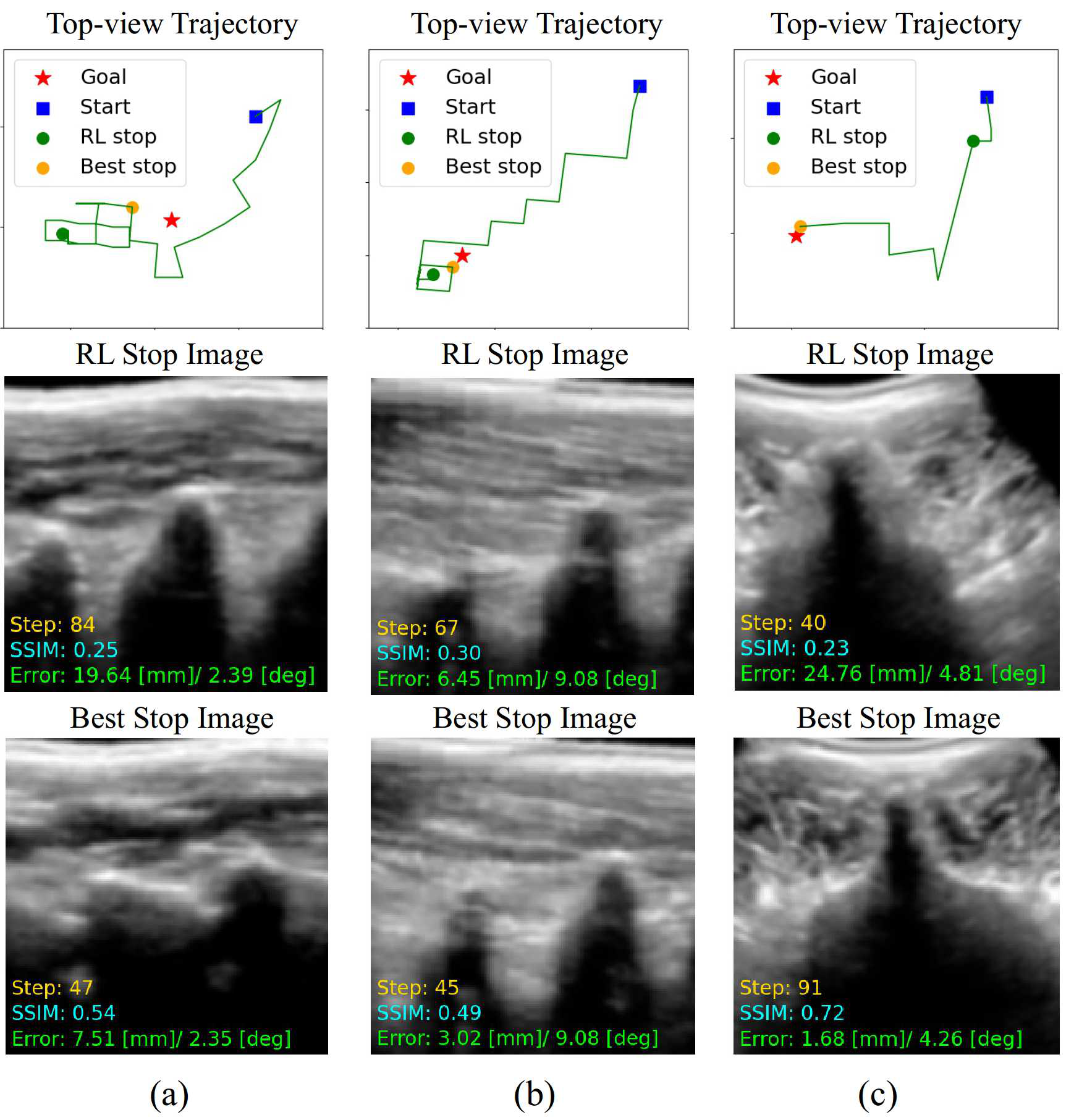}
\caption{(a)-(c) illustrate the PSL, PSAP, and TSP view acquisitions by the ``RL+ASR" and ``RL+ASR+DL" methods on example test cases. The first line shows the top-view navigation trajectories toward each target standard view (green line), with the start and goal positions indicated by a blue square and a red star. The final stopping position of the RL agent (``RL stop") and the best stop determined by dual-agent collaborative navigation (``best stop") are indicated by green and orange balls, respectively. The next two lines show the final images acquired at the RL stop and the best stop, with the corresponding termination step (yellow), final SSIM (cyan) and pose error (green).}
\label{Fig_dual_traj}
\vspace{-0.2cm}  
\end{figure}

We further investigate the effectiveness of the dual-agent collaborative navigation through qualitative analysis. As shown in Fig. \ref{Fig_dual_traj}, the standard view acquisition results of the ``RL+ASR" and ``RL+ASR+DL" methods are compared on example test cases.
It can be observed in Fig. \ref{Fig_dual_traj}(a) that the ``RL+ASR" agent takes a zig-zag path toward the PSL view in the first $\sim40$ steps, and then moves aways from the goal and terminates the navigation at step $84$, ending up with a pose error of $19.64mm/2.39^\circ$ and a low SSIM of $0.25$. While the ``RL+ASR+DL" stops at step $47$, with a pose error of $7.51mm/2.35^\circ$ and an SSIM of $0.54$, showing that using the standard view recognition results for active termination can effectively improve the navigation performance. 
For the PSAP view acquisition, as shown in Fig. \ref{Fig_dual_traj}(b), the ``RL+ASR" agent also gradually approaches the goal at first, but takes some wrong actions in the last $20$ steps and ends up with a slight deviation from the goal ($6.45mm/9.08^\circ$). By applying the dual-agent collaboration navigation method, the stopping position is closer to goal ($3.02mm/9.08^\circ$), and the SSIM is also increased from $0.30$ to $0.49$. 
As shown in Fig. \ref{Fig_dual_traj}(c), during the navigation toward TSP view, the ``RL+ASR" agent gets trapped around the initial position and stops at step $40$ with a large pose error of $24.76mm/4.81^\circ$. In contrast, the ``RL+ASR+DL" agent makes an exploration and begins a new round of search, and successfully reaches the goal at step $91$, with a high accuracy of $1.68mm/4.26^\circ$ and an SSIM of $0.72$. This is because the proposed dual-agent framework can mimic the behavior of a sonographer to randomly reposition the probe on the patient multiple times. As a result, it can escape from bad initial positions to better search for the target anatomy and efficiently navigate the probe toward the goal within the time limit. 

\subsection{Video Demonstration}

Video demonstration is available at \url{https://youtu.be/qhtZ7-vY6M8} and \url{https://www.bilibili.com/video/BV1A3411C78V/} for a better visualization of our results.

\section{Discussion and Conclusion}

In this paper, we have presented a framework that integrates RL and DL techniques for autonomous standard view acquisitions in robotic spinal sonography. The proposed RL agent can automatically control the 6-DOF movement of a US probe based on US images, and utilize the task-specific acoustic shadow information to guide the navigation. The location-sensitive DL agent for standard view recognition can jointly determine the movement of the probe to improve the navigation accuracy and efficiency. Our results in both quantitative and qualitative experiments show the effectiveness of the proposed framework, with an average navigation accuracy of $5.18mm/5.25^\circ$ and $12.87mm/16.97^\circ$ in the intra- and inter-subject settings, respectively. 
It is found that the proposed view-specific ASR and RL-DL collaborative navigation techniques in our framework can effectively improve the navigation performance of a basic RL agent in the US standard view acquisition tasks in both the ``intra-subject" and ``inter-subject" settings. However, it can also be observed in Table~\ref{T_overall} that all the variations of our methods perform worse in the ``inter-subject" setting which requires the agent to interpret anatomical variations among subjects. 
This discrepancy may come from the anatomical differences of the selected subjects in our dataset, which reveals that the current model still has a limited ability to generalize to different patient anatomy. This problem may be mitigated by collecting data from a larger population to build a larger dataset, so that the learned model will be less affected by individual differences and better generalize to different populations. Another possible reason for the discrepancy may lie in the difference in the image quality for different subjects during our data acquisition, as the optimal scanning parameters for different individuals may not be the same. To this end, the scanning parameters (e.g., force applied on the patient) can be tailored to the individual characteristics before the data acquisition to optimize the image quality for each subject \cite{9399640}\cite{virga2016iros}.

In view of the real-world application of the framework, some additional challenges would need to be tackled. First of all, since our method outputs the desired 6-DOF pose of the US probe, which is typically attached to a robot end-effector, the planned motion of the probe needs to be executed using an appropriate controller to ensure safe and compliant motion of the robot. For example, force monitoring and control must be implemented to ensure patient safety and comfort during the robotic acquisition. Also, our simulation of probe-patient interaction has not considered the impact of contact force on the imaging results, such as image distortion caused by tissue deformation.
However, since we focus on the spinal applications, the bone structure can be roughly considered as rigid with little deformation during the scan as long as the force is controlled in a safe range, and the image distortion caused by skin deformation is assumed to be negligible when the robot moves at a relatively low speed. 
In addition, our simulation only considers a static patient during the scan, while patient movement is inevitable in real-world US scans and may cause undesired displacement of the probe. Since we use $1$ DOF of the probe to follow the skin surface (as discussed in Section III-B), an external sensing device such as an RGB-D camera can be used to detect the skin surface in real time \cite{Hennersperger2016MRI} and our algorithm can adjust the probe to the skin surface, thereby overcoming the tissue motions and stabilizing the probe on the skin.
Furthermore, to improve the robustness of the learned policy and better close the gap between simulation and reality for real-world deployment of the method, the simulation environment would need some modifications to be more accurate and realistic (e.g., by introducing additionally recorded force-image paired data and dynamic information of the subjects). 
Other potential directions towards a better sim-to-real transfer include applying robust RL \cite{morimoto2005robust} which explicitly takes into account the disturbances and modeling errors, and incorporating IL techniques \cite{droste2020automatic} to take advantage of expert demonstrations.

Another limitation of the current work is that the neural networks for the RL and DL agents in our framework are separately trained, while they clearly share similarity in the feature extraction stage. In the future work, one may consider a shared network for the two agents that uses the same backbone to extract US image features for both the navigation and classification tasks, thereby reducing the computational cost and memory requirements needed to deploy the framework.
In addition, in this work, we use a sequence of $4$ most recent frames stacked together as the RL agent's observation to take into consideration the historical information. Using other number of stacked frames with frame selection strategies or applying an LSTM \cite{hochreiter1997long} on top of the neural network may make better use of the dynamic information and improve the efficiency of the acquisition, which will need further investigation in our future work.

Despite the challenges that need to be tackled before the method could be used in the clinical setting, our presented work has shown the potential to realize autonomous and intelligent robotic US imaging, and will hopefully pave the way for a promising future of US-based medical care.


%
%
%
%
%
%
\section*{Acknowledgment}
The authors would like to thank Dr. Dongsheng Liu with the Department of Pain, Peking University Shenzhen Hospital, Shenzhen, China, for his help in the data acquisition.
%

\ifCLASSOPTIONcaptionsoff
  \newpage
\fi



%

\bibliographystyle{IEEEtran}  
\bibliography{bare_t-ase_jrnl}

%





\begin{IEEEbiography}[{\includegraphics[width=1in,height=1.25in,clip,keepaspectratio]{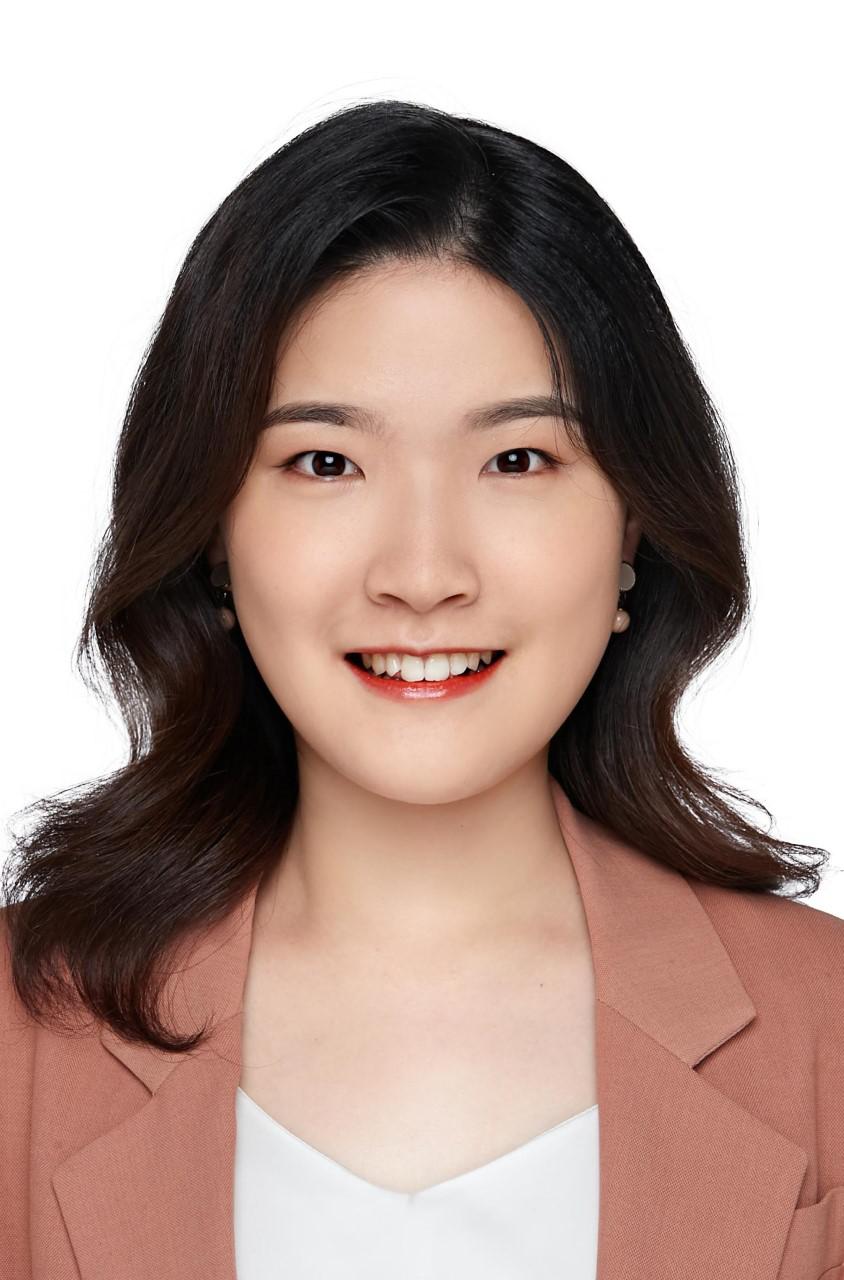}}]
{Keyu Li} received the B.Eng. degree in communication engineering from Harbin Institute of Technology (HIT), Weihai, China, in 2019. She is currently pursuing the Ph.D. degree with the Department of Electronic Engineering, The Chinese University of Hong Kong (CUHK), Hong Kong SAR, China.
Her research interests include artificial intelligence in robot decision-making, medical robotics, and medical imaging applications, with a focus on autonomous robotic ultrasound systems, supervised by Prof. Max Q.-H, Meng.

Ms. Li is an awardee of the Hong Kong PhD Fellowship Scheme (HKPFS) since 2019.
\end{IEEEbiography}
\vfill
\begin{IEEEbiography}[{\includegraphics[width=1in,height=1.25in,clip,keepaspectratio]{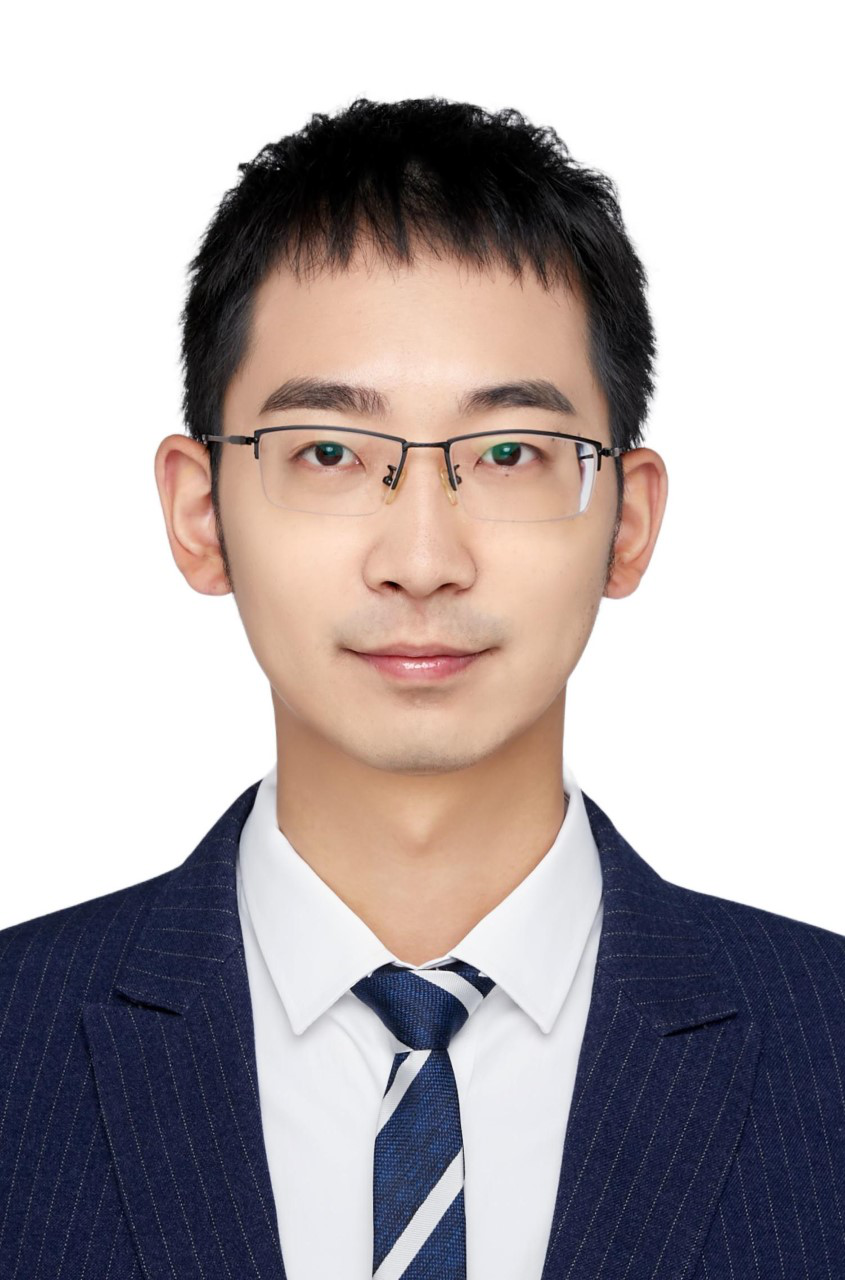}}]{Yangxin Xu}
received the B.Eng. degree in electrical engineering and its automation from Harbin Institute of Technology (HIT), Weihai, Shandong, China, in 2017, and the Ph.D. degree from the Department of Electronic Engineering, The Chinese University of Hong Kong (CUHK), Hong Kong SAR, China, in 2021.
His research focuses on magnetic actuation and localization methods and hardware implementation for active wireless robotic capsule endoscopy, supervised by Prof. Max Q.-H, Meng.

Mr. Xu received the Best Conference Paper from the 2018 IEEE International Conference on Robotics and Biomimetics (ROBIO), Kuala Lumpur, Malaysiain, in 2018.
\end{IEEEbiography}
\vfill

\begin{IEEEbiography}[{\includegraphics[width=1in,height=1.25in,clip,keepaspectratio]{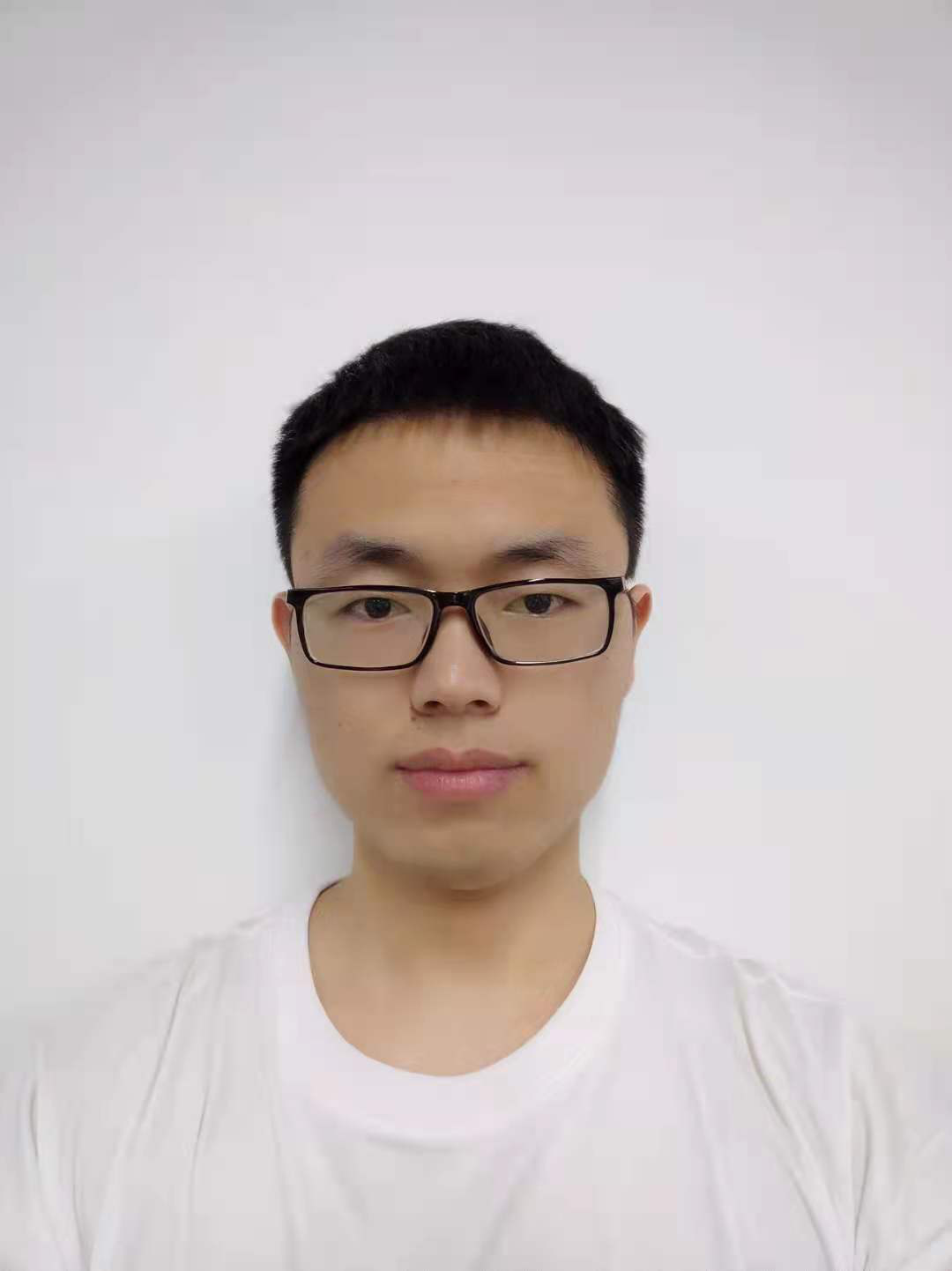}}]{Jian Wang}
received the Master's degree in biomedical engineering from Shenzhen University, China, in 2021. His research interest is in the application of deep learning to ultrasound images, with a particular focus on the breast and thyroid glands. He has published journal papers in Medical Image Analysis and conference papers in MICCAI.
\end{IEEEbiography}

\begin{IEEEbiography}[{\includegraphics[width=1in,height=1.25in,clip,keepaspectratio]{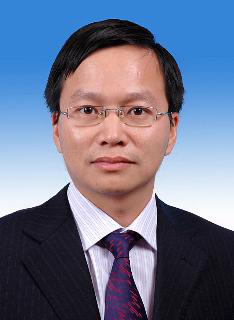}}]{Dong Ni}
received the bachelor's and master's degrees in biomedical engineering from Southeast University, Nanjing, China, in 2000 and 2003, respectively, and the Ph.D. degree in computer science and engineering from The Chinese University of Hong Kong, Hong Kong, in 2009. From 2009 to 2010, he was a Postdoctoral Fellow of the School of Medicine, University of North Carolina at Chapel Hill, USA. Since 2010, he has been with Shenzhen University, China, where he is currently a Professor and the Associate Dean of the Health Science Center, School of Biomedical Engineering. He founded the Medical Ultrasound Image Computing (MUSIC) Laboratory, Shenzhen University. His research interests include ultrasound image analysis, image guided surgery, and pattern recognition.
\end{IEEEbiography}
\vfill

\begin{IEEEbiography}[{\includegraphics[width=1in,height=1.25in,clip,keepaspectratio]{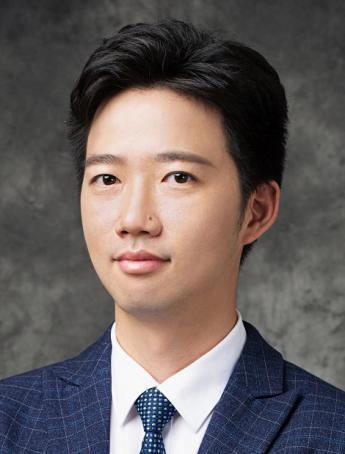}}]{Li Liu}
received the Ph.D. degree in biomedical engineering from the University of Bern, Switzerland, in 2016. He joined as an Assistant Professor with the Health Science Center, School of Biomedical Engineering, Shenzhen University, in 2016. He is currently a Research Assistant Professor with the Department of Electronic Engineering, The Chinese University of Hong Kong. His current research interests include surgical robotics and navigation, ultrasonic/photoacoustic guided surgical intervention, and AI empowered interventions. 

Dr. Liu was a recipient of the Distinguished Doctorate Dissertation Award from the Best Paper Award of IEEE ICIA, in 2009, the MICCAI Student Travel Award, in 2014, and the Swiss Institute of Computer Assisted Surgery, in 2016. He served as a Publication Chair for many international conferences, including a Publication Chair of IEEE ICIA 2017 and 2018 and a Program Chair of IEEE ROBIO 2019. He served as a reviewer for several journals and program.
\end{IEEEbiography}

\vfill

\begin{IEEEbiography}[{\includegraphics[width=1in,height=1.25in,clip,keepaspectratio]{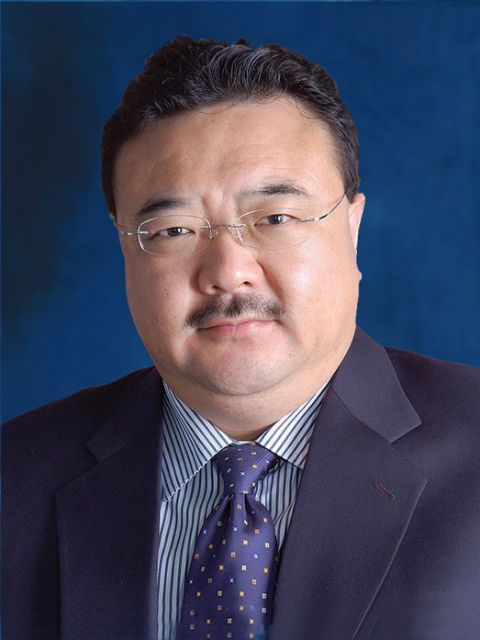}}]
{Max Q.-H. Meng} received the Ph.D. degree in electrical and computer engineering from the University of Victoria, Victoria, BC, Canada, in 1992.

He was with the Department of Electrical and Computer Engineering, University of Alberta, Edmonton, AB, Canada, where he served as the Director of the Advanced Robotics and Teleoperation Laboratory, holding the positions of Assistant Professor, Associate Professor, and Professor in 1994, 1998, and 2000, respectively. In 2001, he joined The Chinese University of Hong Kong, where he served as the Chairman of the Department of Electronic Engineering, holding the position of Professor. He is affiliated with the State Key Laboratory of Robotics and Systems, Harbin Institute of Technology, and is the Honorary Dean of the School of Control Science and Engineering, Shandong University, China. He is currently with the Department of Electronic and Electrical Engineering, Southern University of Science and Technology, on leave from the Department of Electronic Engineering, The Chinese University of Hong Kong, Hong Kong SAR, China, and also with the Shenzhen Research Institute of the Chinese University of Hong Kong, Shenzhen, China. His research interests include robotics, medical robotics and devices, perception, and scenario intelligence. He has published about 600 journal and conference papers and led more than 50 funded research projects to completion as PI.

Dr. Meng is an elected member of the Administrative Committee (AdCom) of the IEEE Robotics and Automation Society. He is a recipient of the IEEE Millennium Medal, a fellow of the Canadian Academy of Engineering, and a fellow of HKIE. He has served as an editor for several journals and also as the General and Program Chair for many conferences, including the General Chair of IROS 2005 and the General Chair of ICRA 2021.
\end{IEEEbiography}

\end{document}